\documentclass[10pt,twocolumn,letterpaper]{article}

\usepackage[pagenumbers]{wacv} 

\usepackage{etex} 
\usepackage{graphicx}
\usepackage{amsmath}
\usepackage{amssymb}
\usepackage{booktabs}
\usepackage{multirow}
\usepackage{subcaption}
\usepackage{enumitem}
\usepackage[pagebackref,breaklinks,colorlinks]{hyperref}
\usepackage{morewrites} 
\usepackage[accsupp]{axessibility}

\usepackage[capitalize]{cleveref}
\crefname{section}{Sec.}{Secs.}
\Crefname{section}{Section}{Sections}
\Crefname{table}{Table}{Tables}
\crefname{table}{Tab.}{Tabs.}

\begin{document}

\title{Dequantization and Color Transfer with Diffusion Models}

\author{
\begin{minipage}{0.3\textwidth}
    \centering
    Vaibhav Vavilala \\
    UIUC \\
    \texttt{vv16@illinois.edu}
\end{minipage}%
\hfill
\begin{minipage}{0.3\textwidth}
    \centering
    Faaris Shaik \\
    UIUC \\
    \texttt{faariss2@illinois.edu}
\end{minipage}%
\hfill
\begin{minipage}{0.3\textwidth}
    \centering
    David Forsyth \\
    UIUC \\
    \texttt{daf@illinois.edu}
\end{minipage}
}

\maketitle

\begin{abstract}
  We demonstrate an image dequantizing diffusion model that enables novel edits on natural images. We propose operating on quantized images because they offer easy abstraction for patch-based edits and palette transfer. In particular, we show that color palettes can make the output of the diffusion model easier to control and interpret. We first establish that existing image restoration methods are not sufficient, such as JPEG noise reduction models. We then demonstrate that our model can generate natural images that respect the color palette the user asked for. For palette transfer, we propose a method based on weighted bipartite matching. We then show that our model generates plausible images even after extreme palette transfers, respecting user query. Our method can optionally condition on the source texture in part or all of the image. In doing so, we overcome a common problem in existing image colorization methods that are unable to produce colors with a different luminance than the input. We evaluate several possibilities for texture conditioning and their trade-offs, including luminance, image gradients, and thresholded gradients, the latter of which performed best in maintaining texture and color control simultaneously. Our method can be usefully extended to another practical edit: recoloring patches of an image while respecting the source texture. Our procedure is supported by several qualitative and quantitative evaluations.
\end{abstract}

\begin{figure}[ht]
\centering
\includegraphics[width=\linewidth]{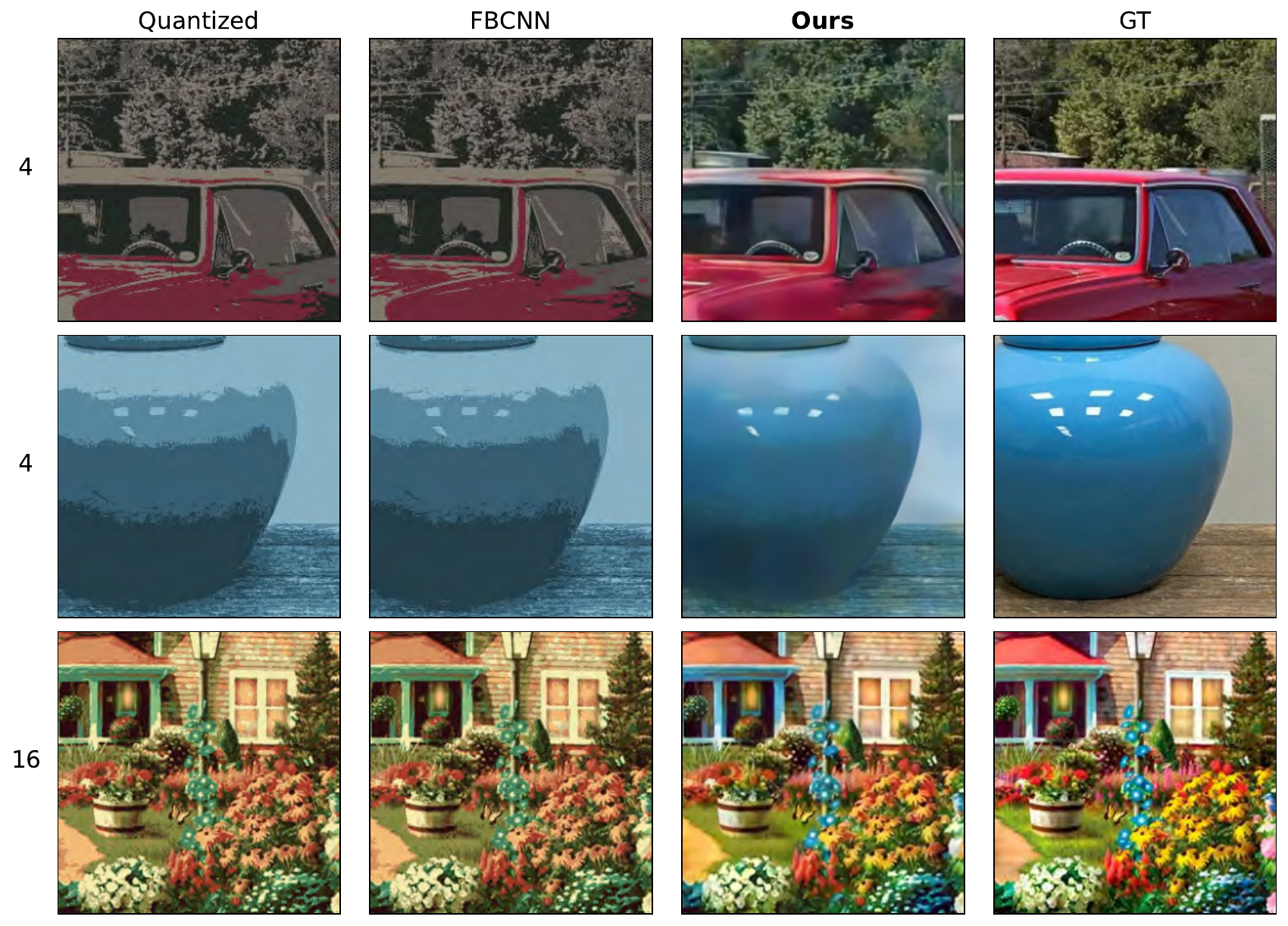}
\caption{We present an image dequantizer that outperforms existing image restoration models. A quantized image is shown in the first column; the second column shows the results of denoising by~\cite{jiang2021towards}; the third is from our procedure (which only requires the quantized image and number of palette colors as input - 4, 16, 64 here); the fourth shows GT. Our model is willing to deviate from the input to produce smoother gradients and life-like colors. Dequantization is not commonly investigated in the image restoration community - here we show why specialized dequantization models are useful, and that existing methods are not sufficient.}
\label{fig:teaser}
\end{figure}

\section{Introduction}
We describe image editing procedures for modifying the color of natural images. Our editing procedures allow artists to apply natural vector art edits to pixel art. For example, an artist can modify the color palette of an image - including the colors in the palette and which patches are assigned to each color. Spatial layout control of reference image based color transfer is not a common capability of existing methods. We use diffusion models to ``snap'' the artist's changes to realistic images.

Diffusion models \cite{dhariwal2021diffusion} have proven to be SOTA in image synthesis. In particular, techniques have been
developed to exert control over the output via text prompts
\cite{nichol2022glide}, edges, segmentation maps, or depth maps
\cite{zhang2023adding}. Some of these procedures do not apply here -- for example, there is no method for editing an image's depth map, and
important and natural controls are not well-explored.

Fine-grained edits to an image involve selecting some portion of the
image, changing it, then obtaining a natural result. Diffusion models
have been shown to accept segmentations as conditioning
(\cite{zhang2023adding,lugmayr2022repaint}), but fine-grained edits have not yet been
shown. Our method accepts a segmentation from any source, such as
~\cite{kirillov2023segment}. A user can then remove one or more patches, and the diffusion model will inpaint the result. Further, the user can specify a color to be used for patches, and the diffusion model will respect that control. This color conditioning is not a standard control in image inpainting (as opposed to text prompts, which are well-established). We argue that changing the color of objects in a scene while maintaining texture is useful in artist workflows. We show that our color \& texture -conditioned inpainting method works well in practice. 

Palette control involves applying a specified palette to a given
image. The palette may come from a pre-built colormap, or more
commonly, an example image. Palette control may involve very large
changes to the color palette, changing the overall feel of the image
(for example, Fig.~\ref{fig:color_transfer}) without disrupting gloss effects, color
gradients, and so on. 

Current style transfer methods (reviewed below) do not apply, because they do not offer spatial layout control over which image patch gets what color. Further, our method offers varying levels of palette abstraction (the number of colors in the palette). Prior methods cannot produce the variety and level of control over the synthesis that our method can. 

For a palette control method to be useful in our application, it
should be able to make very large changes to the palette of the image;
it should be easy to obtain multiple distinct transfers;
and it should be possible to apply detailed edits to the palette
mapping. Our palette control method is built using three tools: vector
quantization of an image's palette; correspondence between color
palettes; and ``vector dequantization,'' where a diffusion model
constructs one or more natural images conditioned on a vector quantized input.
Assume we wish to show image A in B's palette.  We vector quantize
each image's color gamut.  We then build a correspondence between the
centers, then map the colors using the correspondence to
obtain a vector quantized version of the result.  Finally, we vector
dequantize to obtain a natural image.  Choices that affect the result
are: the number of centers chosen in vector quantization;
the particular correspondence process; whether or not to condition on texture (and which method if so); and the randomness
inherent in the diffusion model.

\section{Related Work}
{\bf Image Synthesis:} There is a rich history of image synthesis with GANs
\cite{karras2018progressive,karras2020analyzing,Sauer2023ICML,VavilalaSuperRes,Vavilala_2022_WACV} and
diffusion models
\cite{dhariwal2021diffusion,saharia2022photorealistic}. The quality of
the generated distributions is often quite good, with problems like aliasing
\cite{Karras2021} and sparse datasets \cite{karras2020training}
well-explored. Techniques have been developed to condition the
generation on additional inputs like depth, normals
\cite{esser2021taming,zhang2023adding}, and even 3D primitives~\cite{vavilala2023convex, vavilala2024improvedconvex, vavilala2023blocks}.

{\bf Image inpainting} is a process by which a patch of pixels that has
been removed in an image (say for object removal) must be filled
in (see \cite{elharrouss2020image} for a review). A number of methods
have been developed to solve this problem with CNNs
\cite{liu2018partialinpainting} and diffusion models
\cite{lugmayr2022repaint,xie2023smartbrush}, generating sensible pixels consistent with
the surroundings while preserving texture outside the target region. Masks can be specified to dictate what region to inpaint, and optional text captions can accompany the diffusion model to guide the synthesis. Inpainting is related but different from colorization, which typically takes as input a grayscale image and optionally color hints via keypoints, and synthesizes \textit{ab} channels to accompany an input \textit{L} channel~\cite{zhang2017real,liang2024control}. In contrast, our method accepts a color-quantized image or a natural image with a patch removed for inpainting, with a color hint encompassing the whole patch (improving spatial control). Of critical importance, our method supports luminance changes, whereas existing methods operating in \textit{Lab} space cannot.

We are not aware of an inpainting method that allows the user to specify the approximate color of the inpainted region as an RGB triple with potentially different luminance (previous work only conditions on a text caption e.g. ``red bird,'' or a limited number of colors and objects at low resolution~\cite{khodadadeh2021automatic}). We argue this is an important image editing activity and useful to have in practice. In this work, we show this edit can be accomplished with ControlNet, and a segmenter like SegmentAnything ~\cite{kirillov2023segment,zhang2023adding}; it's likely other inpainting methods can synthesize images conditioned on color, although no one has shown this capability.

{\bf Palette mapping} can be seen as a variant of style transfer,
though we deprecate this view. Style transfer methods
(e.g. \cite{pix2pix2017,kotovenko2019content,luan2017deep,risser2017stable,gatys2015neural,ghiasi2017exploring}) do change palettes, but change the texture of the image as well; in contrast we want to change {\em only} the colors. Some methods require finetuning a style transfer model for each new style~\cite{zhang2023inversionbased,gal2022textual}; others can generalize to new styles~\cite{gatys2015neural,ghiasi2017exploring,Tao2023,deng2022stytr2,jeong2023trainingfree,huang2017arbitrary}. \cite{yang2023zeroshot} uses text prompts rather than images to guide the style transfer.

There is
a small literature of palette transfer methods (with a
review,~\cite{liu2022overview}). Chang {\em et al.} demonstrate the
value of quantizing a palette in a wholly interactive method.
A number of automatic methods are framed as a warp that matches color
spaces.  Reinhard {\em et al.} match moments in \textit{Lab} color
space~\cite{946629}; they obtain improved transfer by segmenting image
and reference and matching segments. Wu {\em et al} use an explicit
semantic representation to match (so flowers to flowers and grass to
grass, say)~\cite{Wu13}.  Piti\'{e} {\em et al.} warp the color space
using repeated 1D projections and a form of histogram
matching~\cite{1544887}; Hwang {\em et al.} use moving least
squares~\cite{6909823}.  Cho {\em et al.} train a network to apply a
specified palette to a natural image using color
augmentations~\cite{Cho_2017_CVPR_Workshops}.  While palette mapping
could be viewed as a colorization problem (decolor, then colorize with
the specified palette), we are not aware of colorization algorithms
that can be forced to use a restricted palette. Existing style transfer methods typically disrupt texture patterns, as well as change color in uncontrolled fashion. Basioti {\em et al.} de-quantize small face images using a GAN~\cite{basioti2020imagedequantizationusinggenerative}.

\begin{figure}[t]
\centering
\includegraphics[width=\linewidth]{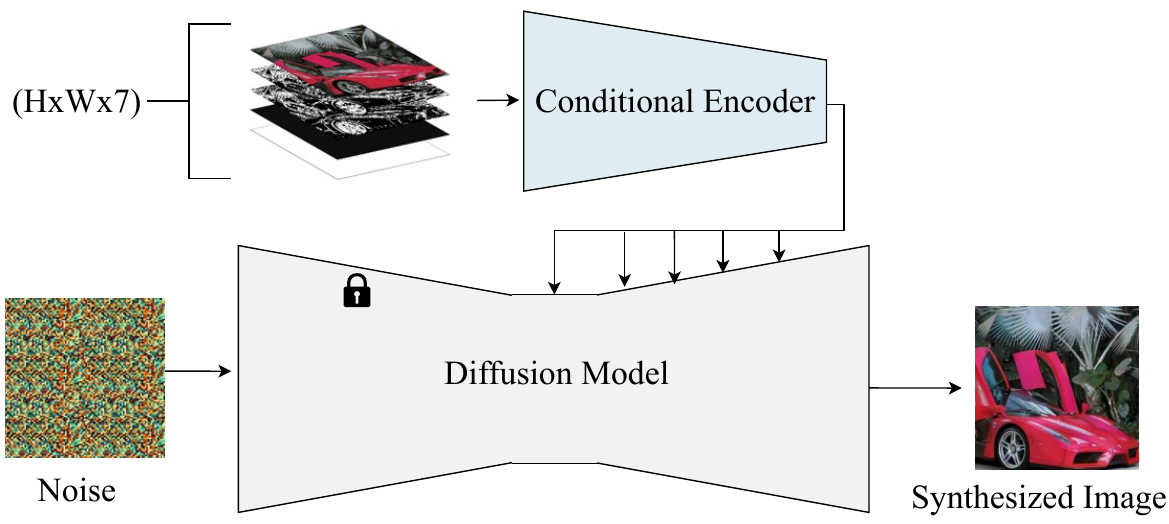}
\caption{Overview of our method. We build a controlled image synthesis pipeline to dequantize images, optionally with texture conditioning. We can apply palette transfer or inpaint patches with a user-specified color, opening new creative applications.}
\label{fig:flow}
\end{figure}

\section{Method}
{\bf Dataset:}
We collect 1.8M images from LAION-Aesthetic\footnote{\url{https://laion.ai/}.} for training, and hold-out a validation and test set. This dataset spans natural images, capturing a wide range of lighting conditions, subjects, and textures. An array of stylistic artwork is represented, such as digital paintings. 

{\bf Architecture:}
Our methodology relies on the principles of ControlNet~\cite{zhang2023adding}, whereby we introduce additional channels to an existing diffusion model. Core to our method is the concept of scale - pretrained diffusion models have seen billions of images, and we'd like to leverage that knowledge. While image restoration models exist for a variety of degradations like deblurring and noise removal, specialized denoisers have been shown to be useful for specific kinds of noise, such as Monte Carlo noise~\cite{vavilala2024denoising}. Here, we aim to remove quantization noise with diffusion models, and show why that is useful. Our foundation model in this work is DeepFloyd Stage II\footnote{\url{https://github.com/deep-floyd/IF}.}~\cite{saharia2022photorealistic}, upon which we build an encoder to accept additional channels as conditioning. As shown in Fig.~\ref{fig:flow}, the base diffusion model is frozen, and the auxiliary encoder module is trainable, extracting useful information from the conditioning and influencing the image synthesis of the base model by adding information to feature maps at varying spatial scales. 

\begin{figure}[t]
    \centering
        \includegraphics[width=0.9\linewidth]{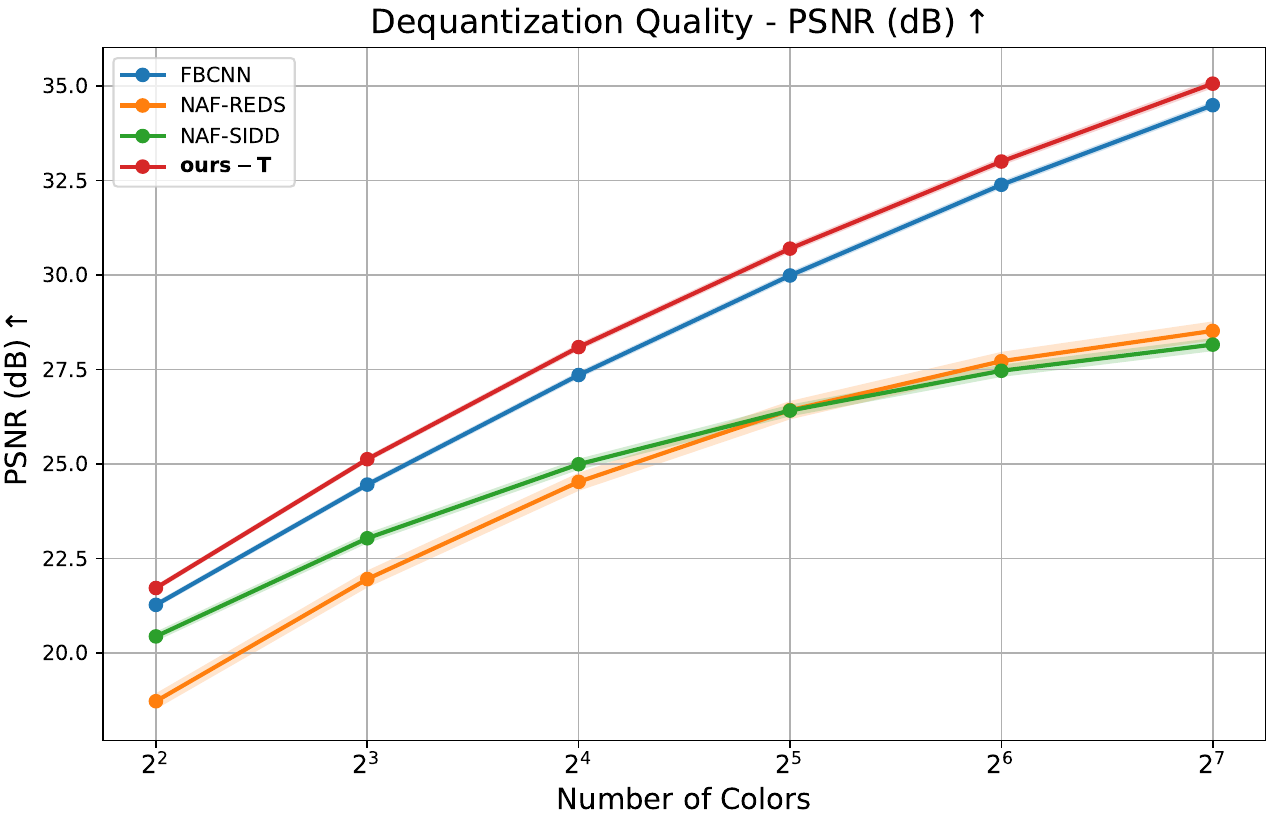}
    \caption{Dequantization without texture conditioning. Our procedure outperforms existing image restoration models across a wide range of palette sizes.}
    \label{fig:recon_noTex}
\end{figure}

{\bf Conditioning Image Dequantizer:}
\textbf{($\mathbf{1^{st}}$ 3 channels))} The primary function of our diffusion model is to dequantize images. Thus, the input to our conditional encoder is the 3-channel RGB image, that contains the quantized image. We use the median cut algorithm from the PIL library to quantize, selecting the number of colors to be powers of 2 from $N \in \{2^2, 2^3, ..., 2^7\}$. \textbf{($\mathbf{4^{th}}$ channel)} We wish to tell the network the quantization level since we know it in advance - we add a fourth channel and let every value be a scalar that takes on the value $N/256$. Optionally, we wish to condition on texture, such that a user can preserve source texture when applying a palette transfer or colorization. Conditioning on luminance is the most natural choice, since existing colorization work uses this channel as conditioning. \textbf{($\mathbf{5^{th} \& 6^{th}}$ channels)} Therefore we train one variant ($\mathsf{ours-L}$) where the fifth channel is the luminance of the GT image at training time. Optionally, at inference time we can replace the luminance of the generated image with the source luminance - we call this configuration $\mathsf{ours-L-post}$ and note that it doesn't warrant training a separate model. One can observe that this form of texture conditioning limits the extent of palette transfers (say for example, a user wants to change a white purse to black while preserving texture; luminance conditioning will prevent the user from getting what they asked for). Thus we train another variant - image gradient conditioning, $\mathsf{ours-G}$. We allocate the fifth and sixth channels of the encoder input to be the image gradients of the luminance channel in the x- and y-directions respectively. While this conditioning modestly improved the range of color transfers we could achieve, the sign and magnitude of the image gradient still encodes local luminance information, which in practice limits the success of extreme palette transfers. Consequently, we train a fourth variant, thresholded gradient: $\mathsf{ours-T}$. We simply threshold the absolute value of the image gradients, resulting in binary images in the fifth and sixth channels consumed by our trainable encoder. $L(i,j) > 8$ worked well in practice, where $L$ is the luminance of the input. As our evaluations show, at the cost of a small amount of texture accuracy, we can better decorrelate local luminance, texture, and color with the thresholded gradient approach. To evaluate our dequantization approach, we compare with recent image restoration methods (see Fig.~\ref{fig:recon_noTex}), designed to remove JPEG artifacts and general image noise~\cite{chen2022simple, jiang2021towards}. Evaluations with these works are done in the regime where texture is not supplied to the model, for fair comparison. \textbf{($\mathbf{7^{th}}$ channel)} We introduce a seventh channel, a binary texture indicator, to tell the network if texture is available at a particular pixel $i,j$. Thus we are training three independent networks, $\mathsf{L(\theta; \textbf{c}),G(\theta; \textbf{c}),T(\theta; \textbf{c})}$ with identical settings except the texture conditioning. $\theta$ refers to the parameters of the network, including the frozen base diffusion model and trainable encoder; $\textbf{c}$ refers to the 7 channel auxiliary conditions described above. Diffusion models are also conditioned on the time step $\textbf{t}$ and text prompt (we use the empty string).

{\bf Inference - Palette Transfer:} Once we establish that our method can effectively dequantize, then we can use color palettes as an effective abstraction for image editing. For example, we can edit one or more colors in the palette, or which patches get assigned which color. Further, our method can dequantize images over a wide range of colors in the palette, supporting varying levels of abstraction. To perform a palette transfer, we can quantize a source image and extract its palette. From there, we select a target color map, which could be extracted from another image. Then we perform weighted bipartite matching, assigning each source color the most similar target color, ending up with a 1:1 correspondence. Given this modified quantized image, we can then dequantize using our diffusion model to obtain the palette-transferred result. The exact palette matching procedure is up to the artist - our focus in this work is to show that our model respects the palette that was asked for, even extreme palette transfers. We show results for two color-matching algorithms. We call matching based on the most similar color as the $\mathtt{color}$ method; by negating the pairwise distance matrix between source and target colors, we can effectively match based on \textit{most dissimilar color}, which results in interesting effects ($\mathtt{negative-color}$).

{\bf Color-conditioned inpainting:}
Our model naturally lends itself to solve a problem closely related to image dequantization: image inpainting. In particular, we can condition the missing patch on a color of the user's choosing, and optionally preserve source texture when inpainting the patch, which is not an edit discussed in the inpainting literature~\cite{saharia2022palette}. To support color-conditioned inpainting, we modify the training process of the existing three networks as follows. We supply the GT image for the first 3 channels, but randomly replace a patch of the image with the mean color of its support. We set the color indicator channel to 1 where GT color is available, and $1/256$ for the masked out patch. Similarly, we can randomly set the texture channels to 0 within the mask. Thus each foundation model that we trained, $\mathsf{L, G, T}$, is capable of dequantizing a wide range of images and colors in the palette as well as performing color-conditioned inpainting - including respecting or ignoring source texture based on user request. As we show in our experiments, even though the network is supplied with the mean color within the patch during training, at inference time any color can be supplied, even if it would appear strange in a natural image, and our method respects what was asked for. 

{\bf Training Details:}
We train our three models for 2 epochs each on a single A40 GPU, requiring approx. five days. For both inpainting and dequantization, the network is trained to predict the GT image given the conditioning, using standard losses from~\cite{saharia2022photorealistic}. We use AdamW optimzer, mixed-precision training, batch size 12, and learning rate $0.00001$. Despite the limited amount of training, we achieve highly competitive and useful results. Even though our model was trained on 256-res crops, it supports any aspect ratio at inference time. Inference at 256-res requires approx. 3 seconds on an A40 GPU; inference at 1920$\times$1080 requires approx. 63 seconds. We found that the $super27$ DDPM sampler performs well, and we experimentally set the dynamic thresholding $p$ and $c$ hyperparameters to 0.95 and 1.0 respectively. We remark that our early experimentation showed that our approach works just as well on latent diffusion models like Stable Diffusion~\cite{esser2021taming}.

\begin{figure}[ht]
    \centering
    \begin{minipage}[b]{\linewidth}
        \centering
        \includegraphics[width=0.9\linewidth]{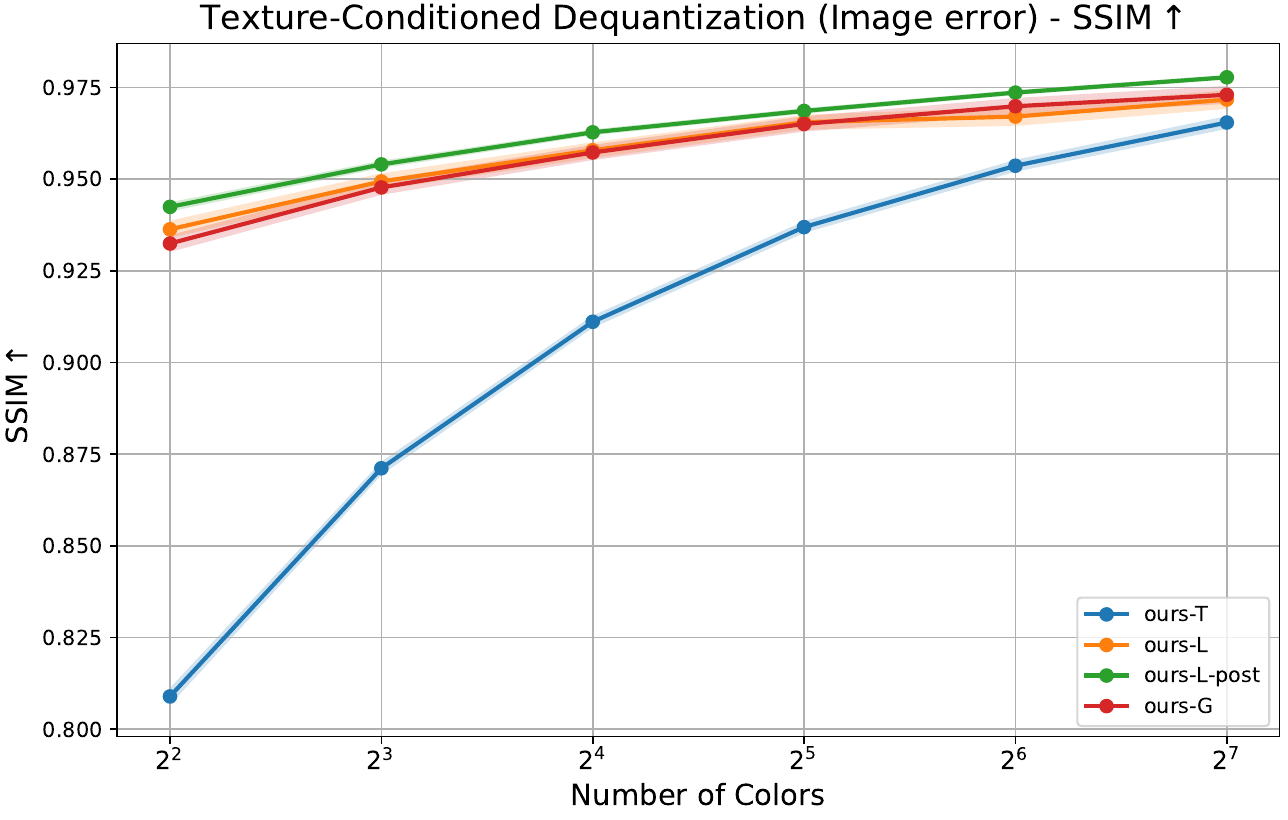}

        \label{fig:recon_a}
    \end{minipage}
    \vfill
    \begin{minipage}[b]{\linewidth}
        \centering
        \includegraphics[width=0.9\linewidth]{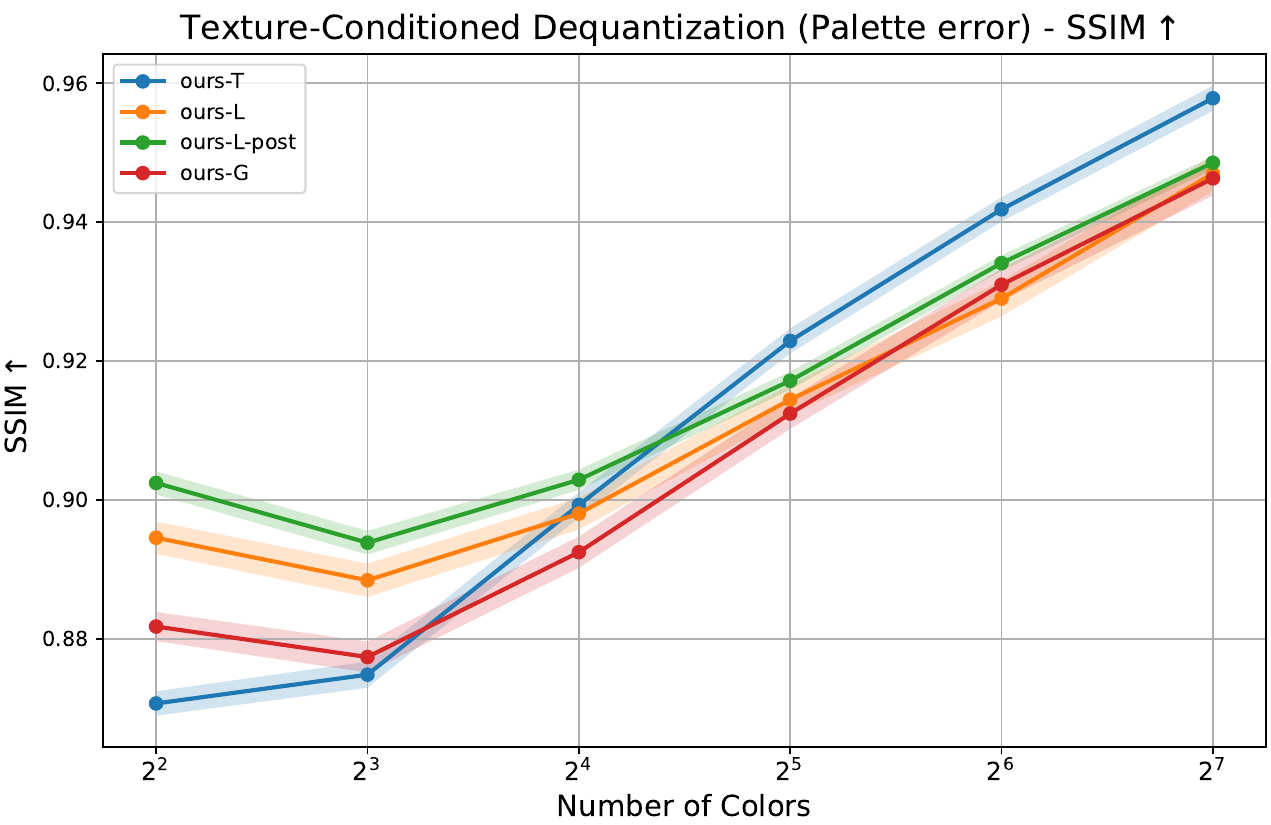}

        \label{fig:recon_b}
    \end{minipage}
    \caption{Dequantization with texture conditioning. \textbf{Top} All methods can dequantize fairly well, though $\mathsf{ours-T}$ suffers in reconstruction quality. \textbf{Bottom} When measuring error w.r.t. to quantized output vs. quantized input, all methods generally excel, though the fact that $\mathsf{ours-T}$ holds the least information and performs better in some cases indicates imperfections in the quantization algorithm (we use median-cut).}
    \label{fig:eval_tex_recon}
\end{figure}

\begin{figure*}[h!]
\centering
\includegraphics[width=0.93\linewidth]{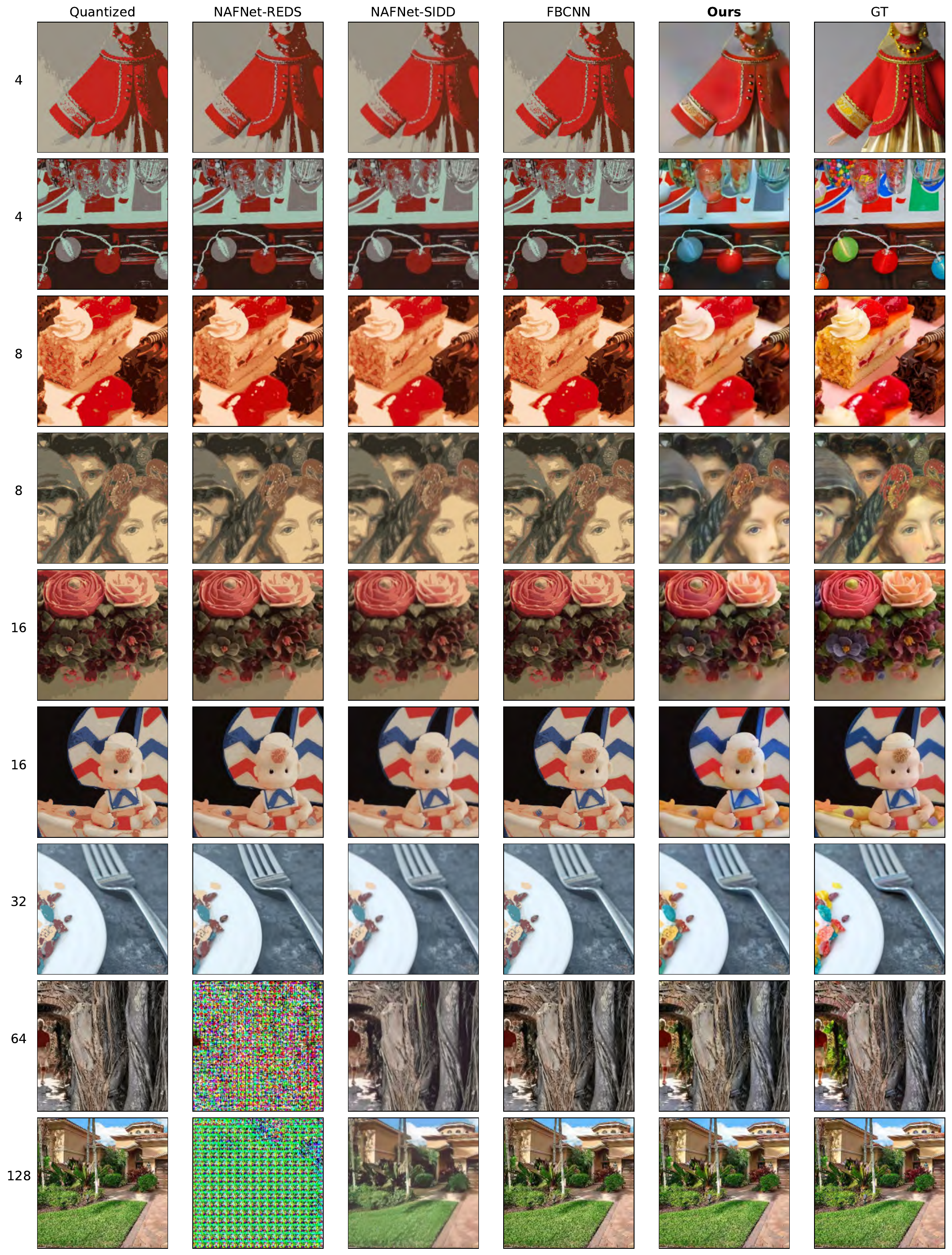}
\caption{Dequantization without texture conditioning - only the quantized image is the input (with number of colors specified in the row). Across the board, our method (\textbf{fifth column}) produces more natural and aesthetic results as compared with baseline denoisers~\cite{chen2022simple,jiang2021towards}.}
\label{fig:noTex_qual}
\end{figure*}

\section{Results \& Discussion}
{\bf Dequantization Quality:}
Core to our method is measuring how well our diffusion models can dequantize natural images. To evaluate, we quantize 1000 test images at several palette sizes, and measure how well our model can predict the source GT image. As shown in Fig.~\ref{fig:recon_noTex}, our model outperforms all baselines, which suggests a specialized dequantization model is warranted. Shading indicates standard error bars. Fig.~\ref{fig:noTex_qual} shows qualitative evaluation. Our method is able to smooth harsh quantization boundaries and restore desaturated colors to a natural level. Our method succeeds because it was specifically trained to restore quantized images; existing denoisers are able to restore JPEG noise but not the aggressive decimation from quantization.

\begin{figure*}[ht]
\centering
\includegraphics[width=0.86\linewidth]{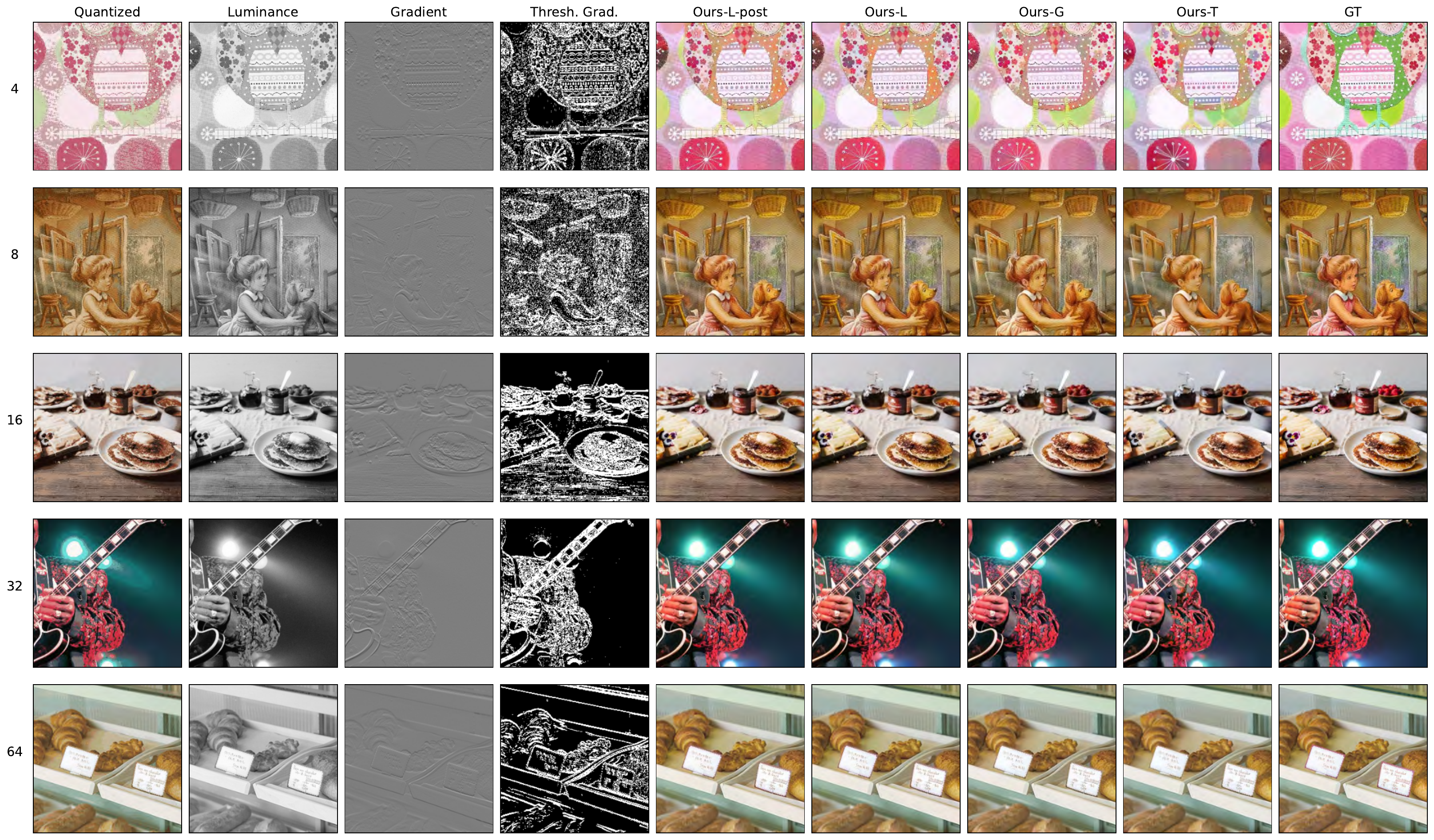}
\caption{Qualitative evaluation of dequantization when texture is available. All methods are generally successful, though methods with luminance available ($\mathsf{ours-L-post}$, $\mathsf{ours-L}$) naturally perform better.}
\label{fig:recon_tex}
\end{figure*}

{\bf Dequantization with Texture:}
In Fig.~\ref{fig:eval_tex_recon}, we evaluate our models when texture conditioning is available. {\bf(top)} We are quantizing the GT image and attempting to reconstruct it from its palette, along with additional texture conditioning (i.e. thresholded gradient, gradient, luminance). {\bf(bottom)} In the case of palette transfers, we do not know the GT image; thus to evaluate, we can quantize the generated image and compare it (i.e. calculate the SSIM/PSNR) with the input (which is a quantized image). Observe how the quality improves with more colors in the palette. Thus, the Palette-SSIM metric is a reasonable proxy when GT is not available, but it's not perfect. Notice that $\mathsf{ours-T}$ (thresholded gradient) performs better than $\mathsf{ours-L-post}$ (which conditions on GT luminance and enforces it as post-process). Palette error has curious difficulties that follow from the behavior of the quantization algorithm. If we impose the original image’s luminance, we might change the quantization substantially. In Fig.~\ref{fig:recon_tex} we evaluate our four methods when texture accompanies the quantized image as conditioning. The row indicates the number of colors in the palette; the first column is the quantized image. The subsequent 3 columns show conditioning texture for the 3 models. $\mathsf{ours-L-post}$ and $\mathsf{ours-L}$ condition on luminance, with the former enforcing luminance as a post-process. $\mathsf{ours-G}$ conditions on image gradients and $\mathsf{ours-T}$ conditions on thresholded gradients. All models - even those with less information available - are capable of dequantizing the input quite well, though the luminance-based models tend to achieve better color reproduction.

\begin{figure*}[ht]
\centering
\includegraphics[width=\linewidth]{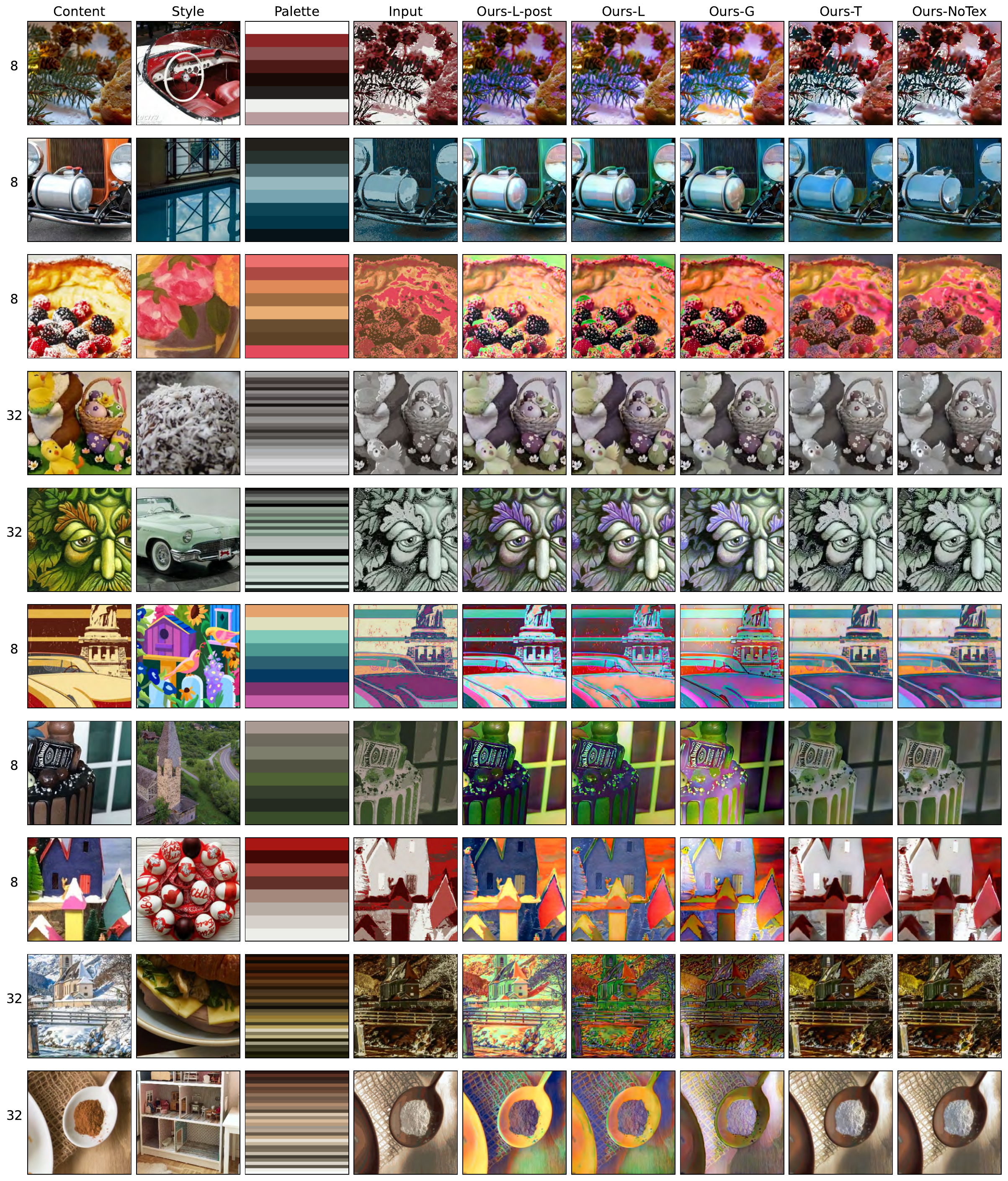}
\caption{Using color palettes as a form of abstraction, we can transfer color from one image ($\mathbf{2^{nd} column}$) to another ($\mathbf{1^{st} column}$). The extracted palette is shown in the $\mathbf{3^{rd} column}$, and the quantized input to our method is shown in the $\mathbf{4^{th} column}$. The remaining five columns show the synthesized results. Models with a weaker notion of texture (last two columns) can better respect the requested palette.}
\label{fig:color_transfer}
\end{figure*}

{\bf Palette Transfer:}
A key motivation for good image dequantization is palette based image editing. In Table~\ref{tab:color_palette_metrics}, we evaluate using our palette error metric (because GT is not available). The palette is transferred from a random test image via minimum bipartite graph matching for selecting colors, where most similar color is the criteria ($\mathtt{color}$ method). Note that error metrics are generally worse with texture conditioning - this is because with texture, the network must find solutions that respect both the palette and texture that was asked for, which results in deviations from the palette. When the palette is the only conditioning, it's easier to respect it. Thresholded gradient ($\mathsf{ours-T}$) generalizes the best when texture is supplied. All methods succeed when texture is not supplied, because the conditioning is identical amongst the models in those settings (except $\mathsf{ours-L-post}$, where luminance is not an input but is still enforced as a post-process). We present qualitative palette transfer results in Fig.~\ref{fig:color_transfer}. Five rows of $\mathtt{color}$ method and five rows of $\mathtt{negative-color}$ are shown. $\mathsf{ours-T}$ excels in generating natural results that respect both the texture and requested palette, because the texture conditioning is better decorrelated with the palette conditioning.

\begin{figure*}[ht]
\centering
\includegraphics[width=0.85\linewidth]{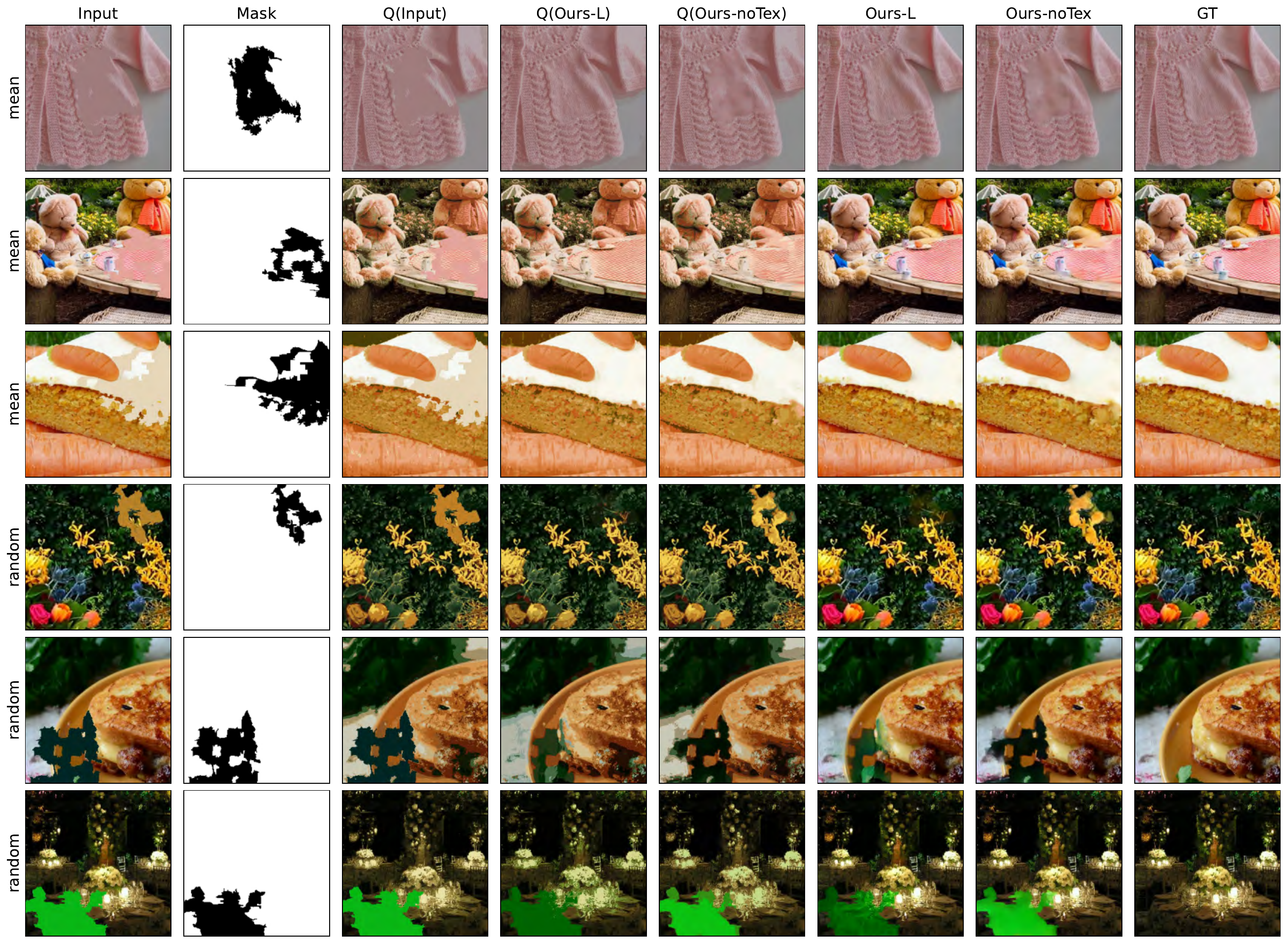}
\caption{Our image dequantizer can inpaint with color and/or texture conditioning, a useful artist edit.}
\label{fig:patch_main}
\end{figure*}

{\bf Inpainting:}
Because our method relies on a pretrained diffusion model, and it can map quantized image patches to natural images, we can present another application: color-conditioned image inpainting. We show qualitative results in Fig~\ref{fig:patch_main}. The first column shows the input with a patch recolored with: (first 3 rows) the mean color of its support  or (bottom 3 rows) a random color. The Mask column shows the affected region in black. The next three columns show the result of quantizing: the input; our luminance-conditioned result; and our no-texture-conditioned result. $\mathsf{ours-L}$ and $\mathsf{ours-noTex}$ are generated results. The final column is the ground truth image. In the first three rows, our method respects the requested color and texture fairly well, though the absence of texture (col. 7) causes some smoothing. Row four shows the problem with luminance conditioning: $\mathsf{ours-L}$ is incapable of generating a bright yellow color since that is different from the dark luminance in that patch. The texture and color conditioning conflict with each other. Rows 5 and 6 show similar outcomes. Generally, luminance conditioning results in pleasing and well-harmonized images that do not always respect the exact color that was asked for. Removing texture conditioning trades-off local shading reproduction and harmonization in exchange for color accuracy. In the absence of ground truth recolored images (final 3 rows), measuring palette error between $\textbf{Q(Ours)}$ and $\textbf{Q(Input)}$ provides an indication of whether the user got what they asked for.

{\bf Conclusion:}
We have presented an image dequantization procedure and demonstrated it can perform extreme palette transfers. We proposed several foundation models with slightly different conditioning and trade-offs. In practice, all methods should be available in an artist's toolbox.

\begin{table}[t!] {
\centering
\setlength{\tabcolsep}{1pt} 
\resizebox{\linewidth}{!}{
\begin{tabular}{l||cc|cc|cc|cc}
\hline
\multirow{3}{*}{Method} & \multicolumn{4}{c|}{Texture On} & \multicolumn{4}{c}{Texture Off} \\
 & \multicolumn{2}{c}{8 colors} & \multicolumn{2}{c|}{32 colors} & \multicolumn{2}{c}{8 colors} & \multicolumn{2}{c}{32 colors} \\
 & PSNR$\uparrow$ & SSIM$\uparrow$ & PSNR$\uparrow$ & SSIM$\uparrow$ & PSNR$\uparrow$ & SSIM$\uparrow$ & PSNR$\uparrow$ & SSIM$\uparrow$ \\
\hline
$\mathsf{ours-T}$ & \textbf{23.0} & \textbf{0.774} & \textbf{27.4} & \textbf{0.874} & 25.2 & \textbf{0.842} & 28.8 & 0.901 \\
$\mathsf{ours-G}$ & 17.8 & 0.516 & 19.1 & 0.538 & 25.6 & 0.834 & 28.5 & 0.893 \\
$\mathsf{ours-L}$ & 16.3 & 0.526 & 15.8 & 0.490 & \textbf{25.7} & 0.839 & \textbf{29.1} & \textbf{0.905} \\
$\mathsf{ours-L-post}$ & 13.2 & 0.481 & 12.8 & 0.444 & 12.8 & 0.468 & 12.6 & 0.457 \\
\hline
\end{tabular}}}
\vspace{-1em}
\caption{Palette transfer metrics for $\mathtt{color}$  method. Even though $\mathsf{ours-T}$ performed slightly worse in texture-conditioned dequantization (Fig.~\ref{fig:eval_tex_recon}), it excels in palette transfers because thresholded gradients decorrelate local texture from local luminance.}
\vspace{-2em}
\label{tab:color_palette_metrics}
\end{table}

{\small
\bibliographystyle{ieee_fullname}
\bibliography{egbib}
}

\clearpage  
\twocolumn[
\begin{center}
    \section*{Supplement - Dequantization and Color Transfer with Diffusion Models}
\end{center}
]
\addcontentsline{toc}{section}{Supplement}

\section{Additional Color Transfer Results}
\begin{figure*}[ht]
\centering
\includegraphics[width=\linewidth]{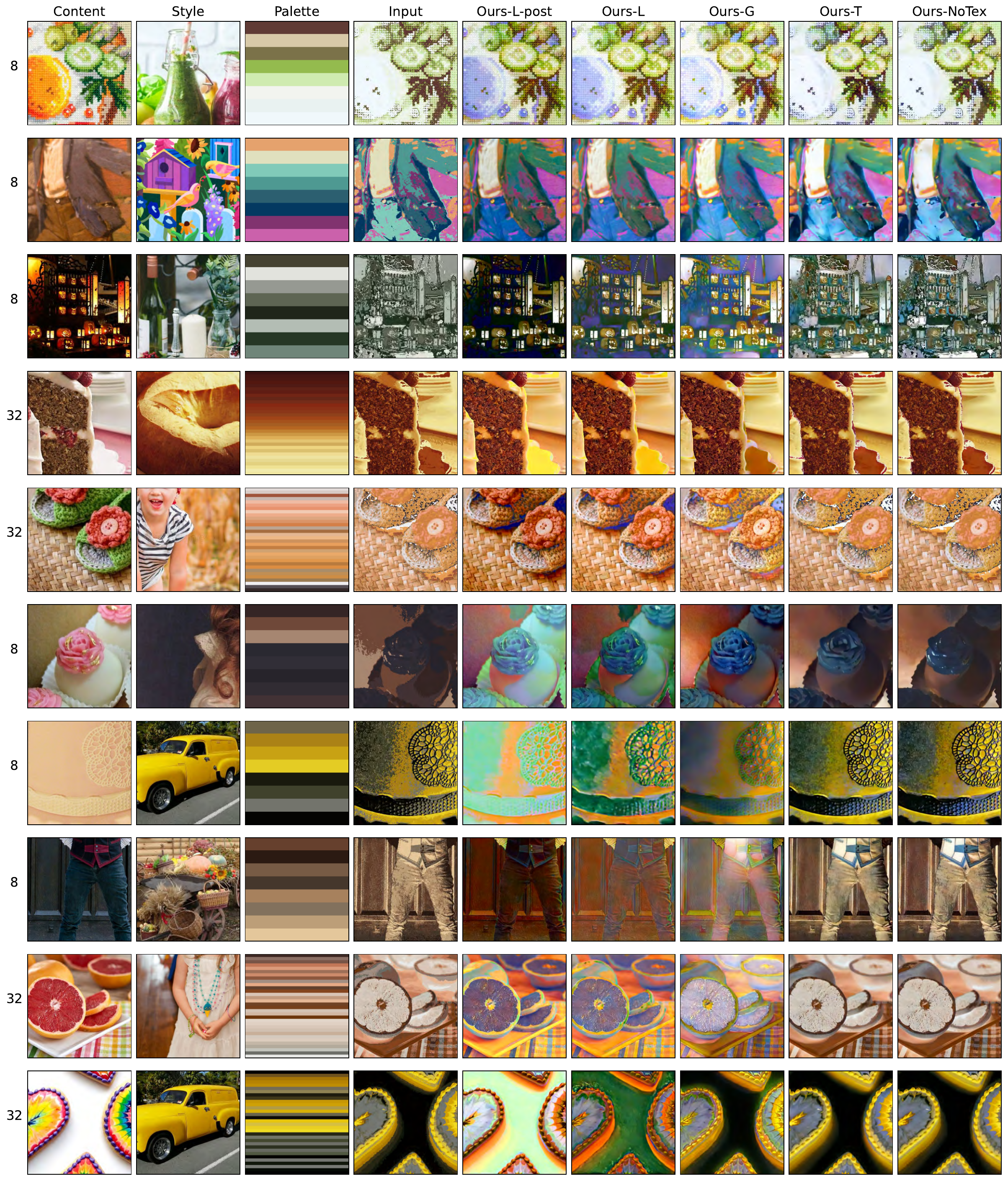}
\caption{Additional palette transfer samples.}
\label{fig:color_transfer_supp}
\end{figure*}

\begin{figure*}[ht]
\centering
\includegraphics[width=\linewidth]{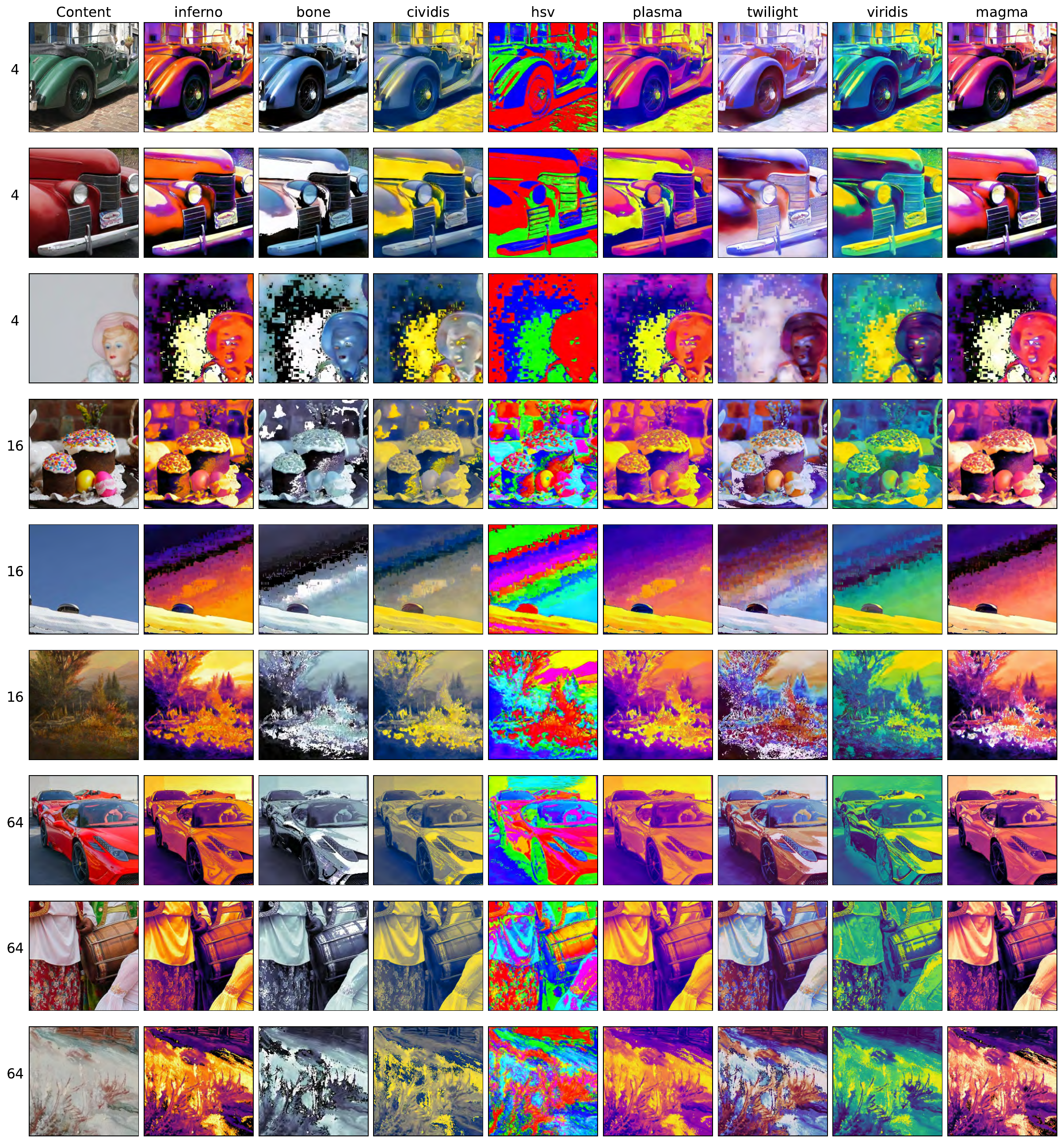}
\caption{Colormap transfer examples for $\mathsf{ours-T}$. The row indicates the number of colors in the palette when performing the transfer from a real matplotlib color map (see column name). Our generative model is conditioned on the transferred color palette and texture, and dequantized results are shown. Our method opens a number of creative possibilities.}
\label{fig:cmap_T}
\end{figure*}

\begin{figure*}[ht]
\centering
\includegraphics[width=\linewidth]{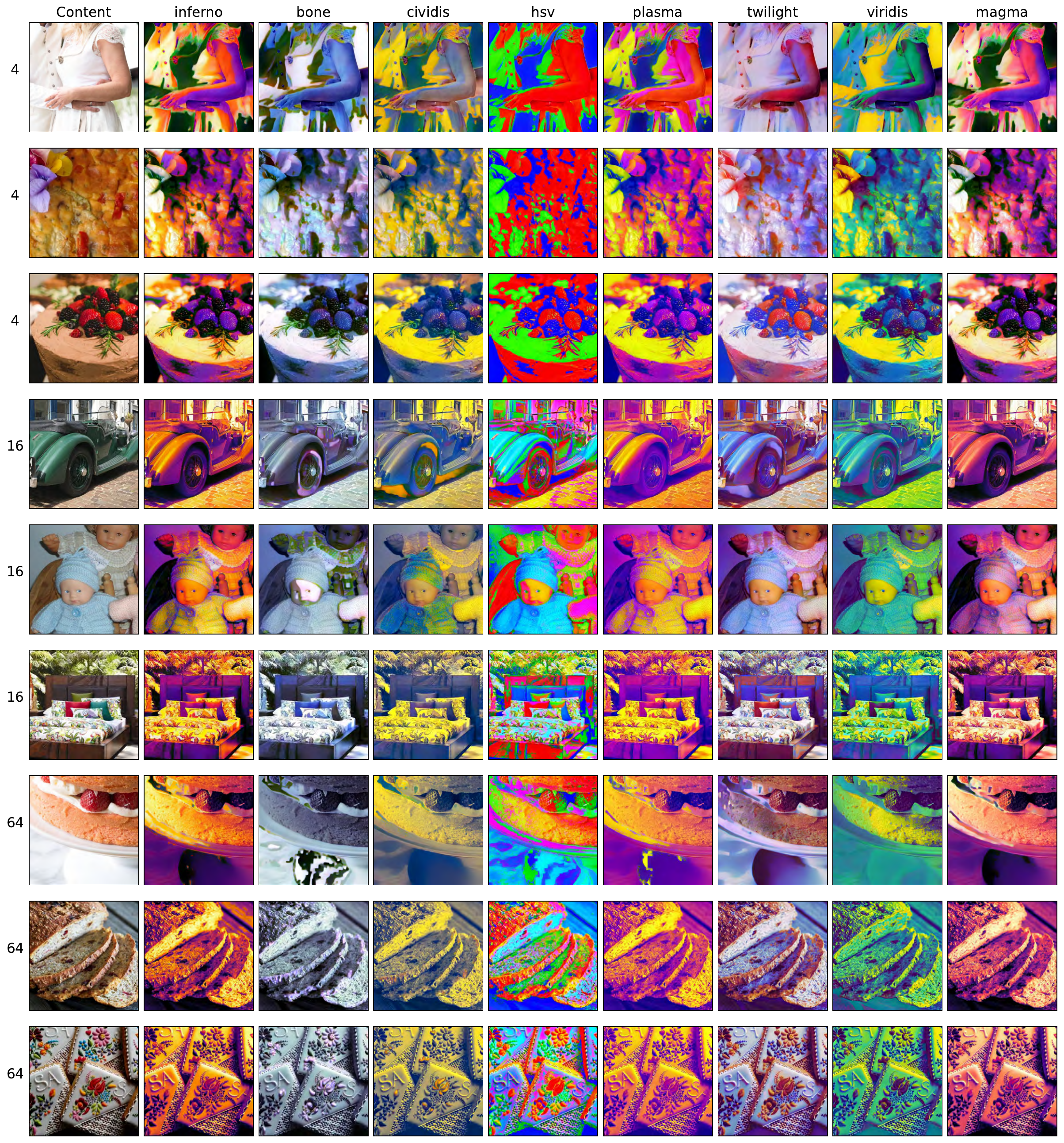}
\caption{Colormap transfer examples for $\mathsf{ours-G}$. The row indicates the number of colors in the palette when performing the transfer from a real matplotlib color map (see column name). Our generative model is conditioned on the transferred color palette and texture, and dequantized results are shown. Our method opens a number of creative possibilities.}
\label{fig:cmap_G}
\end{figure*}

\begin{figure*}[ht]
\centering
\includegraphics[width=\linewidth]{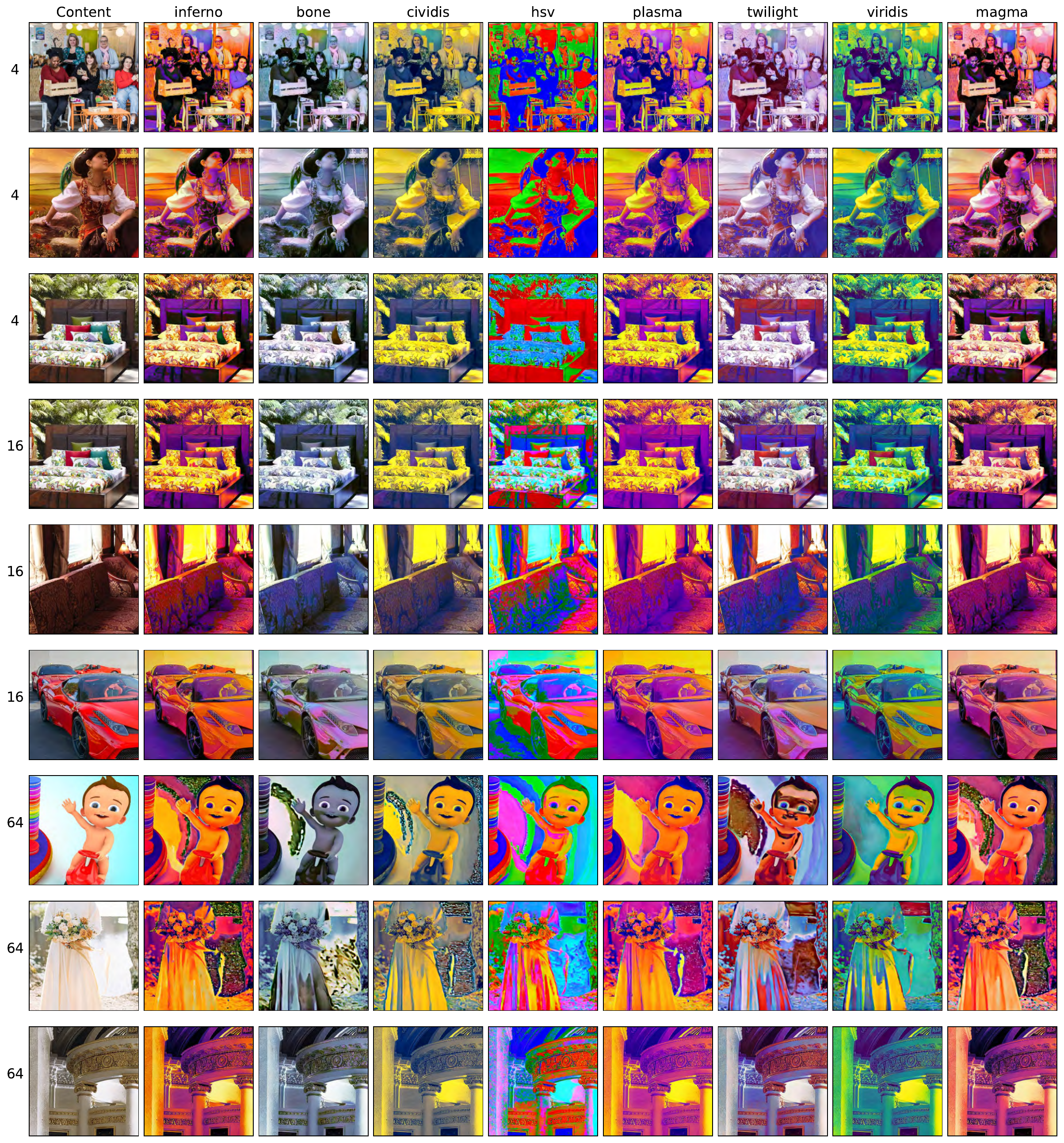}
\caption{Colormap transfer examples for $\mathsf{ours-L}$. The row indicates the number of colors in the palette when performing the transfer from a real matplotlib color map (see column name). Our generative model is conditioned on the transferred color palette and texture, and dequantized results are shown. Our method opens a number of creative possibilities.}
\label{fig:cmap_L}
\end{figure*}

\begin{figure*}[ht]
\centering
\includegraphics[width=\linewidth]{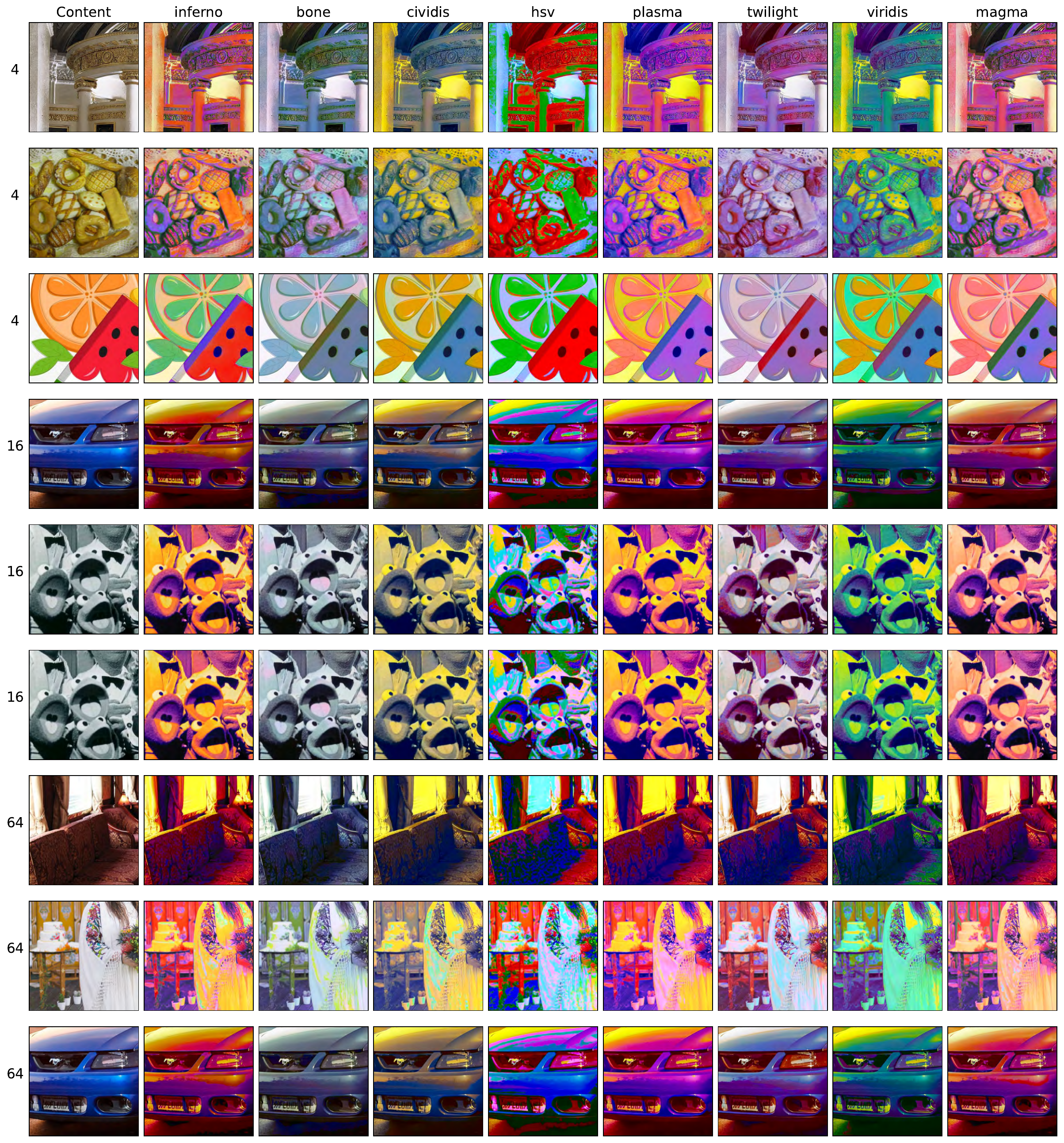}
\caption{Colormap transfer examples for $\mathsf{ours-L-post}$. The row indicates the number of colors in the palette when performing the transfer from a real matplotlib color map (see column name). Our generative model is conditioned on the transferred color palette and texture, and dequantized results are shown. Our method opens a number of creative possibilities.}
\label{fig:cmap_Lpost}
\end{figure*}

In Table~\ref{tab:neg_palette_metrics}, we quantitatively evaluate the $\mathtt{negative-color}$ method, examining the palette error because GT is not available. We present additional qualitative color transfer results in Fig.~\ref{fig:color_transfer_supp}, in which we transfer color palettes from real images. We can even transfer color palettes from pre-built color maps, obtaining pleasing results. We show results from each model (using the $\mathtt{color}$ method) in Figs.~\ref{fig:cmap_T},~\ref{fig:cmap_G},~\ref{fig:cmap_L},~\ref{fig:cmap_Lpost}. 

\begin{table}[h] {
\centering
\setlength{\tabcolsep}{1pt} 
\resizebox{\linewidth}{!}{
\begin{tabular}{l||cc|cc|cc|cc}
\hline
\multirow{3}{*}{Method} & \multicolumn{4}{c|}{Texture On} & \multicolumn{4}{c}{Texture Off} \\
 & \multicolumn{2}{c}{8 colors} & \multicolumn{2}{c|}{32 colors} & \multicolumn{2}{c}{8 colors} & \multicolumn{2}{c}{32 colors} \\
 & PSNR$\uparrow$ & SSIM$\uparrow$ & PSNR$\uparrow$ & SSIM$\uparrow$ & PSNR$\uparrow$ & SSIM$\uparrow$ & PSNR$\uparrow$ & SSIM$\uparrow$ \\
\hline
$\mathsf{ours-T}$ & \textbf{24.4} & \textbf{0.784} & \textbf{27.7} & \textbf{0.851} & 25.8 & \textbf{0.824} & 28.7 & 0.873 \\
$\mathsf{ours-G}$ & 13.8 & 0.107 & 16.0 & 0.107 & 25.7 & 0.807 & 28.4 & 0.864 \\
$\mathsf{ours-L}$ & 10.4 & 0.0991 & 10.2 & 0.0410 & \textbf{25.9} & 0.814 & \textbf{29.0} & \textbf{0.880} \\
$\mathsf{ours-L-post}$ & 7.64 & 0.0177 & 7.37 & -0.0362 & 6.99 & -0.0588 & 6.81 & -0.0853 \\
\hline
\end{tabular}}}
\caption{Palette transfer error metrics for $\mathtt{negative-color}$ method. This metric measures error between the quantized output and quantized input, because GT is not available. \textbf{Texture On} We condition each method on texture and the quantized source image after performing an extreme palette transfer. For gradient and luminance conditioning, the signal from the palette may indicate a different local luminance than the signal from the texture, preventing the method from synthesizing the color that was asked for. Note the lower PSNR for these methods, as compared with $\mathsf{ours-T}$. Thresholded gradient is the best choice for performing extreme palette transfers while preserving texture. \textbf{Texture Off} The conditioning in this case is identical amongst the models - only the quantized source image (post color transfer) and the number of colors in the palette. In the absence of texture conditioning, all methods perform comparably well, except $\mathsf{ours-L-post}$, which still enforces the $L$ channel onto the generated image.}
\label{tab:neg_palette_metrics}
\end{table}

\section{Additional Patch Editing Examples}
In a practical artist workflow, an artist may be working with an image but want to edit the color of a patch or object, while possibly keeping its texture the same. Our dequantization models are naturally capable of performing this task, whereby the network accepts the target image with a patch of pixels replaced by the requested color, and the network will fill in the missing region while loosely following the requested color. In our evaluations, we show our models can perform this task quite well. 

\textbf{Quantitative evaluation} We select 1000 random test images at 256-res, and pick random patches in each to remove. On average, $13.4\%$ of pixels are masked out. In Table~\ref{tab:patch_mean_err}, we replace the removed patch with the mean color of its support, and measure how well the model can reconstruct the image. As expected, more texture information is helpful, as shown by the strong numbers for $\mathsf{ours-L}$. However, evaluating each model's ability to support extreme color changes tells a different story. In Table~\ref{tab:patch_rand_err}, we repeat this experiment, but we replace each patch with a random $(R,G,B)$ triple. Here, conditioning on texture resulted in the thresholded gradient model performing best, while luminance-conditioning performed worst. This is because conditioning on luminance prevented the network from generating colors with a different luminance that what was asked for in the patch conditioning. Note that in the absence of GT, we use palette error as a proxy error metric. This metric is not perfect, as for example the models without texture conditioning performed slightly better than with texture. However, the palette error gives a strong indication as to whether the model generated an image whose color palette matches what was requested. We use 16 colors in the palette when running this error metric. 

\begin{figure*}[ht]
\centering
\includegraphics[width=\linewidth]{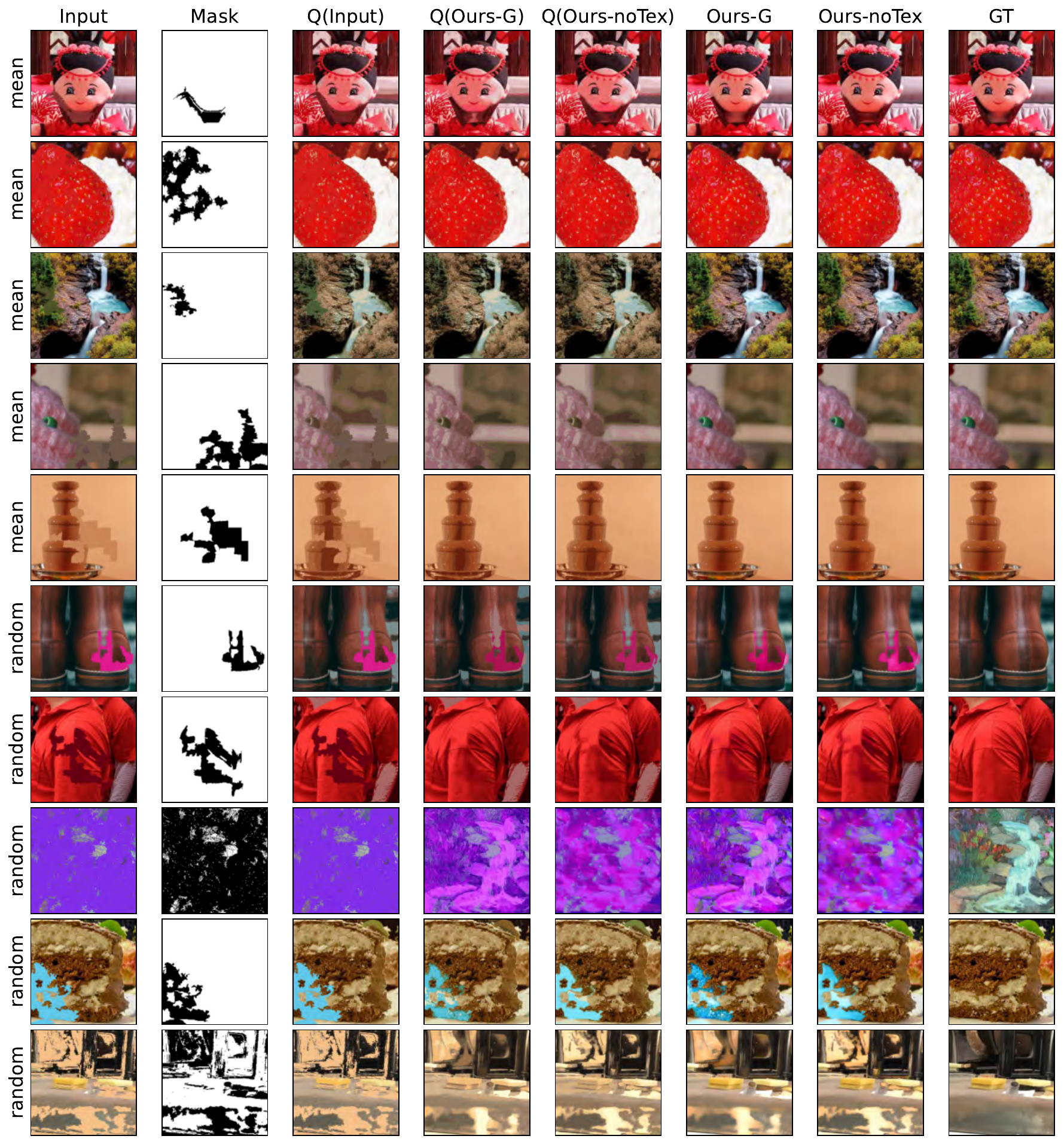}
\caption{Additional patch editing samples from $\mathsf{ours-G}$.}
\label{fig:patch_G_supp}
\end{figure*}

\begin{figure*}[ht]
\centering
\includegraphics[width=\linewidth]{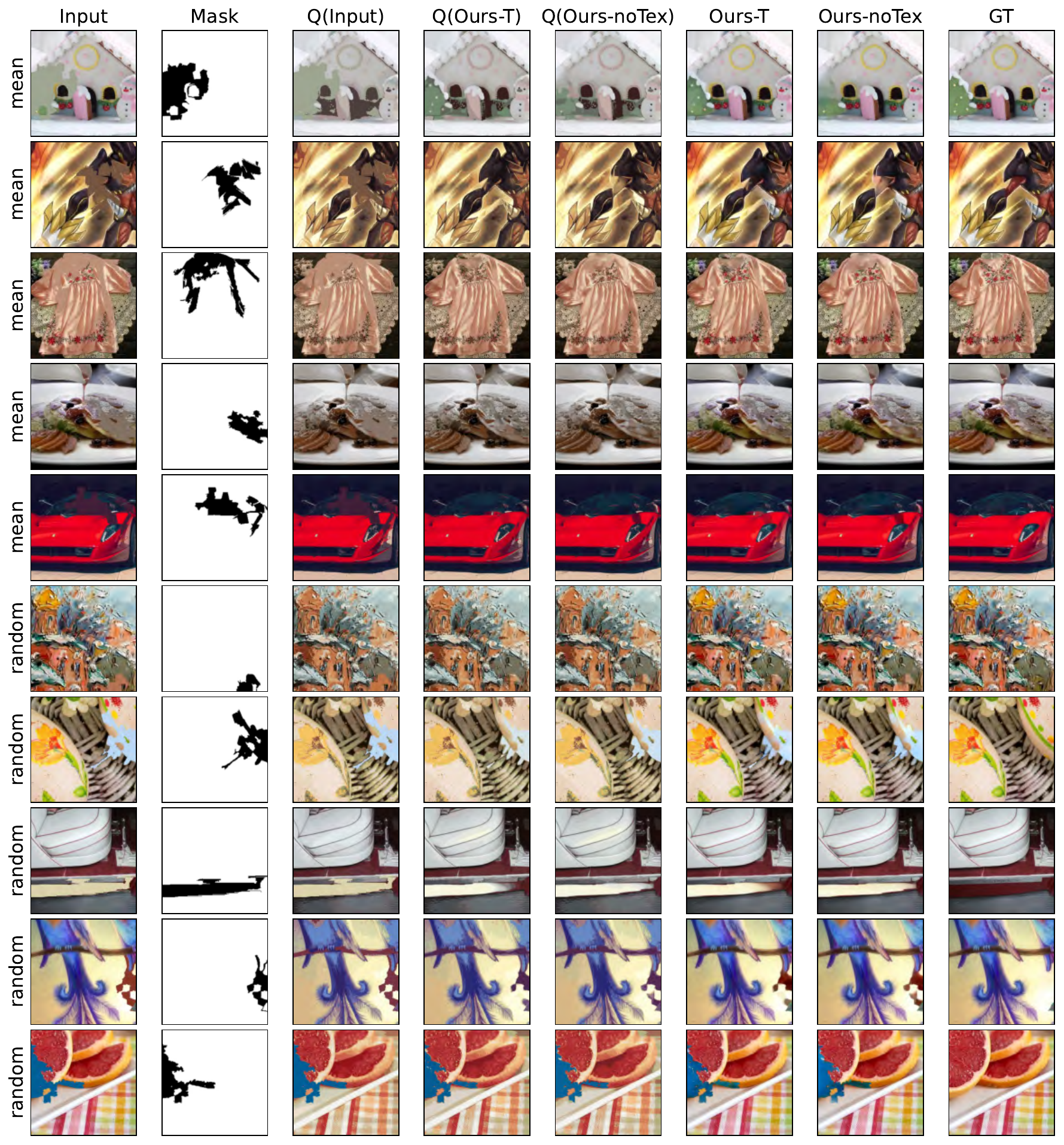}
\caption{Additional patch editing samples from $\mathsf{ours-T}$.}
\label{fig:patch_T_supp}
\end{figure*}

\begin{figure*}[ht]
\centering
\includegraphics[width=\linewidth]{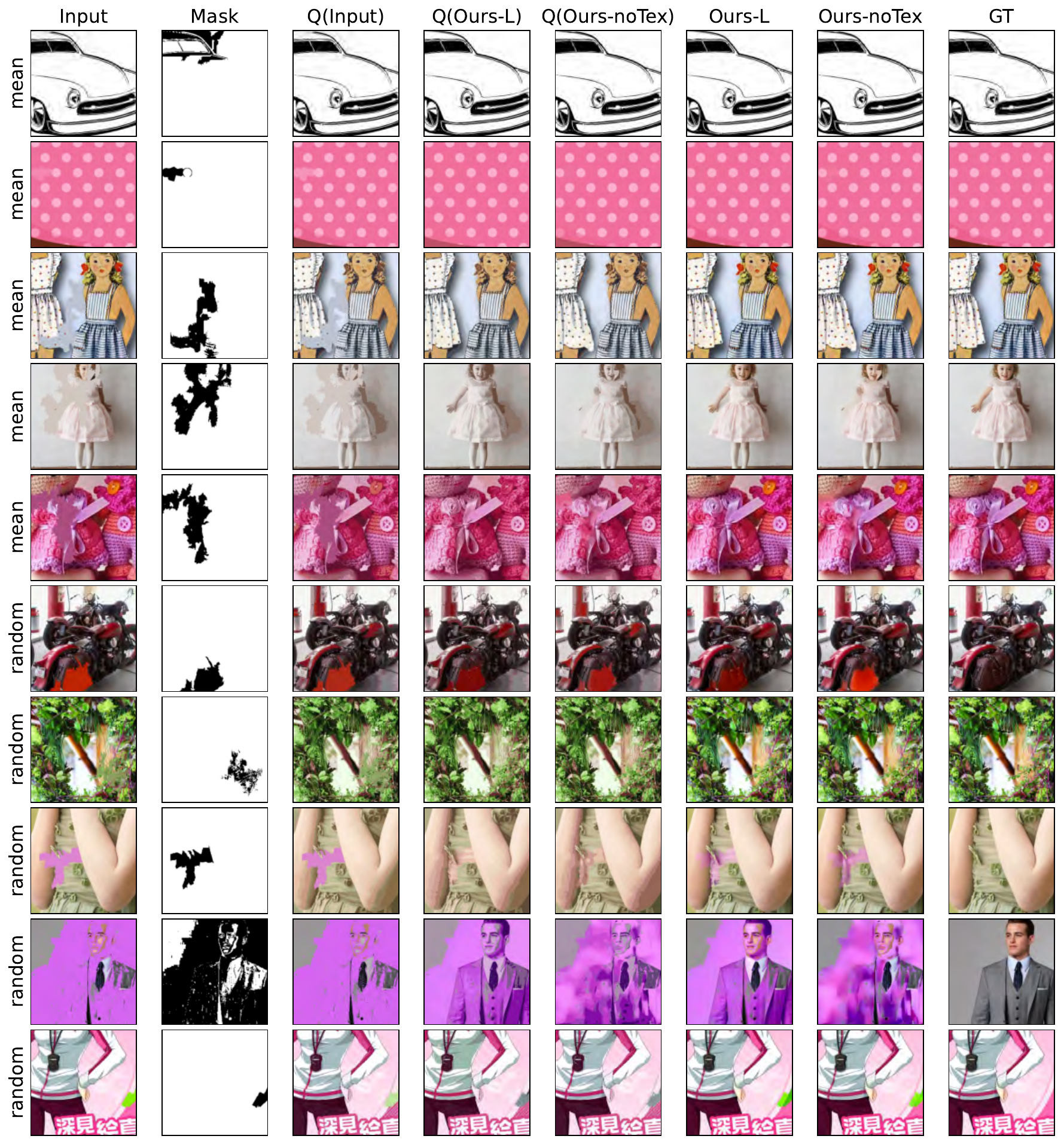}
\caption{Additional patch editing samples from $\mathsf{ours-L}$.}
\label{fig:patch_L_supp}
\end{figure*}

\textbf{Qualitative evaluation} We present additional patch editing examples using our dequantization models in Figs.~\ref{fig:patch_G_supp},~\ref{fig:patch_T_supp}, and~\ref{fig:patch_L_supp}. Color-conditioned inpainting is not a well-explored area within the inpainting literature - our family of models supports this capability. For small color changes, $\mathsf{ours-L}$ is the best choice; extreme color changes look best with our $\mathsf{ours-T}$, because thresholded gradients can better decorrelate local luminance from texture. $\mathsf{ours-G}$ is a reasonable mid-ground. Ultimately, it's up to the artist but our analysis provides a practical starting point.

\begin{table}[t]
\centering
\resizebox{\linewidth}{!}{%
\begin{tabular}{l|cc|cc}
\hline
\multirow{2}{*}{Inpainting quality - mean} & \multicolumn{2}{c|}{Texture On} & \multicolumn{2}{c}{Texture Off} \\
 & PSNR$\uparrow$ & SSIM$\uparrow$ & PSNR$\uparrow$ & SSIM$\uparrow$ \\
\hline
$\mathsf{ours-T}$ & 30.68 & 0.9336 & 29.3 & 0.921 \\
$\mathsf{ours-G}$ & 38.77 & 0.9823 & 30.95 & 0.9315 \\
$\mathsf{ours-L}$ & \textbf{40.12} & 0.9845 & 29.68 & 0.9202 \\
$\mathsf{ours-L-post}$ & 39.82 & \textbf{0.9867} & \textbf{38.27} & \textbf{0.9834} \\
\hline
\end{tabular}%
}
\caption{Image reconstruction error metrics when the selected patch is replaced by the mean color within its support. Error is measured between generated image and ground truth. \textbf{Texture On} Conditioning on texture is clearly better than not; and using the luminance performs best, followed by gradient and thresholded gradient. Interestingly, enforcing the luminance as a post-process ($\mathsf{ours-L-post}$) performed marginally worse than not doing so ($\mathsf{ours-L}$). \textbf{Texture Off} The three base models all perform similarly because the conditioning is identical - quantized image only without texture. $\mathsf{ours-L-post}$ outperforms the other methods because it still enforces the \textit{L}-channel. This table illustrates the loss in reconstruction quality when reducing the strength of texture conditioning. This is valuable to know, because we gain color control in exchange for weaker texture control, a possibly acceptable tradeoff depending on the image editing use-case.}
\label{tab:patch_mean_err}
\end{table}

\begin{table}[h]
\centering
\resizebox{\linewidth}{!}{%
\begin{tabular}{l|cc|cc}
\hline
\multirow{2}{*}{Inpainting quality - random} & \multicolumn{2}{c|}{Texture On} & \multicolumn{2}{c}{Texture Off} \\
 & PSNR$\uparrow$ & SSIM$\uparrow$ & PSNR$\uparrow$ & SSIM$\uparrow$ \\
\hline
$\mathsf{ours-T}$ & \textbf{23.56} & \textbf{0.8241} & 23.87 & 0.8285 \\
$\mathsf{ours-G}$ & 22.37 & 0.7964 & 23.58 & 0.8208 \\
$\mathsf{ours-L}$ & 21.76 & 0.7895 & \textbf{24.07} & \textbf{0.8347} \\
$\mathsf{ours-L-post}$ & 20.57 & 0.7739 & 20.53 & 0.7731 \\
\hline
\end{tabular}%
}
\caption{Palette error metrics when the selected patch is replaced by a random color. Note that GT is not available, so we use palette error as a proxy. \textbf{Texture On} Observe that thresholded gradient performed best, and luminance conditioning worst, because thresholded gradients better decorrelate texture from color. \textbf{Texture Off} The three base models all perform similarly because the conditioning is identical - quantized image only without texture.}
\label{tab:patch_rand_err}
\end{table}

\section{Evaluating Robustness}
\begin{figure}[ht]
\centering
\includegraphics[width=\linewidth]{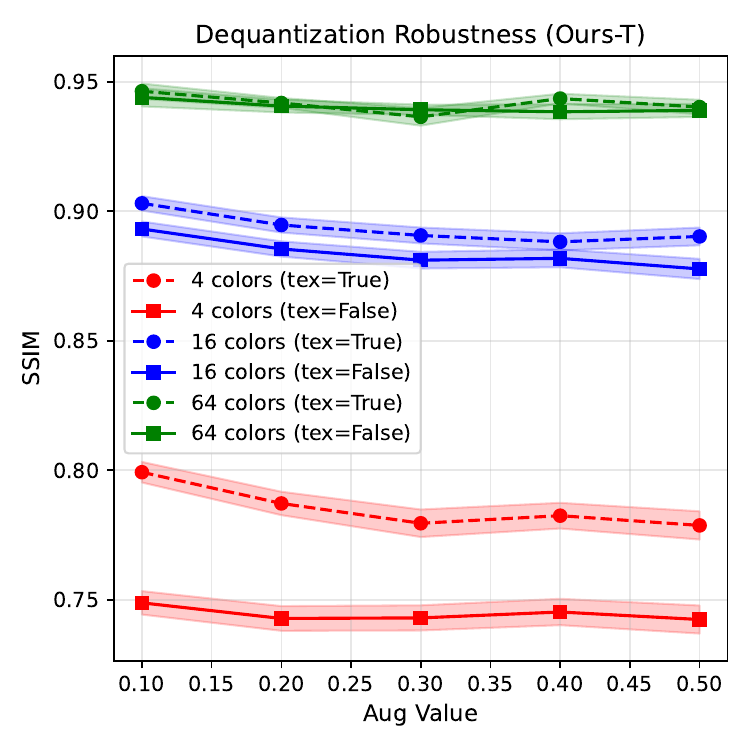}
\caption{Our dequantization method is successful even when aggressively augmenting the source palette. In this case, we can apply the same augmentation to the source palette $Q(I)$ and the GT image $I$, allowing us to compute error metrics between the generated image and $A(I)$, the new GT image. Each data point represents the average SSIM over 300 random test images, with shaded standard error bars.}
\label{fig:deq_sigma}
\end{figure}

In Fig.~\ref{fig:deq_sigma}, we evaluate the robustness of our procedure to increasingly aggressive augmentations, which push the input quantized image outside the training distribution. To evaluate, we pick a random test image $I$, and apply a random augmentation to it, obtaining $A(I)$. The augmentation is in HSV space, with the maximum magnitude of change controlled by $Aug$ $value$, the x-axis. We also quantize the source test image and then apply the same augmentation to it, obtaining $A(Q(I))$, which serves as conditioning to our model $\mathsf{ours-T}$. We optionally condition on thresholded gradient (dotted lines) or no texture conditioning (solid lines). We can then compute error metrics between the generated image, $\mathsf{T(\theta; \textbf{c}=A(Q(I)))}$, and the target image, $A(I)$. The success of our method across a sweep of augmentation strengths suggests that the quantized image is an effective abstraction of the original image, as changes to the palette are reflected in the model output. Further, our model can generate images from strange palettes quite different from the training set, as reflected by the success of our method at higher augmentation levels. Observe how texture conditioning is more important when fewer colors are available in the palette, as more details are abstracted away from the quantized image in these settings. The fact that our base diffusion model has seen billions of images is a huge advantage in maintaining robustness.

\clearpage

\begin{figure*}[ht]
\centering
\includegraphics[width=\linewidth]{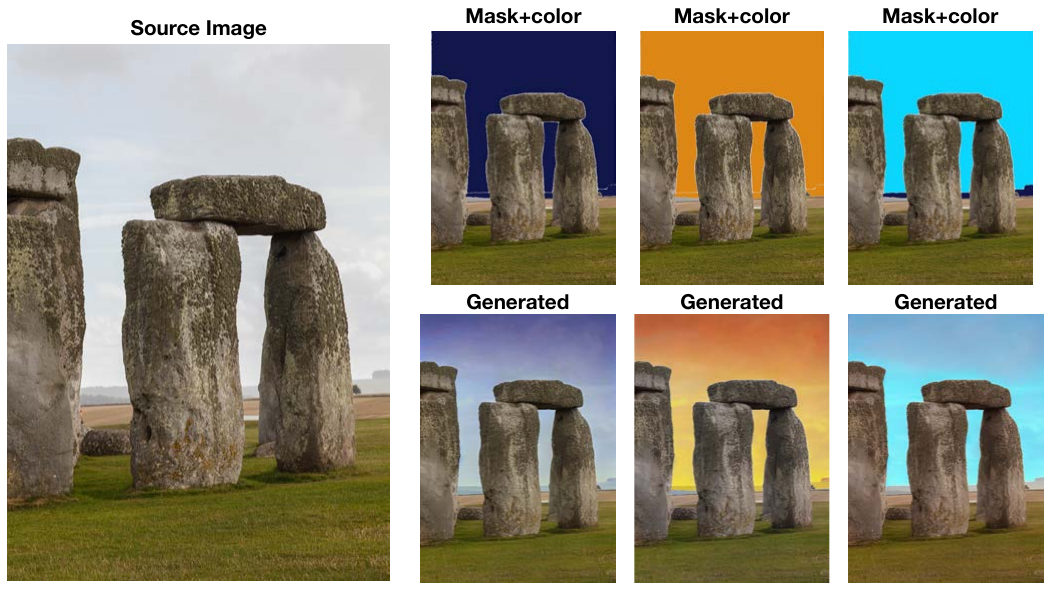}
\caption{Practical color-conditioned inpainting example. We obtain the sky (via selecting a Segment Anything patch~\cite{kirillov2023segment}), pick colors of interest, and have the conditional diffusion model dequantize (we use gradient-based conditioning here). The model harmonizes the result fairly well, respecting the requested color and mask.}
\label{fig:stone}
\end{figure*}

\begin{figure*}[ht]
\centering
\includegraphics[width=\linewidth]{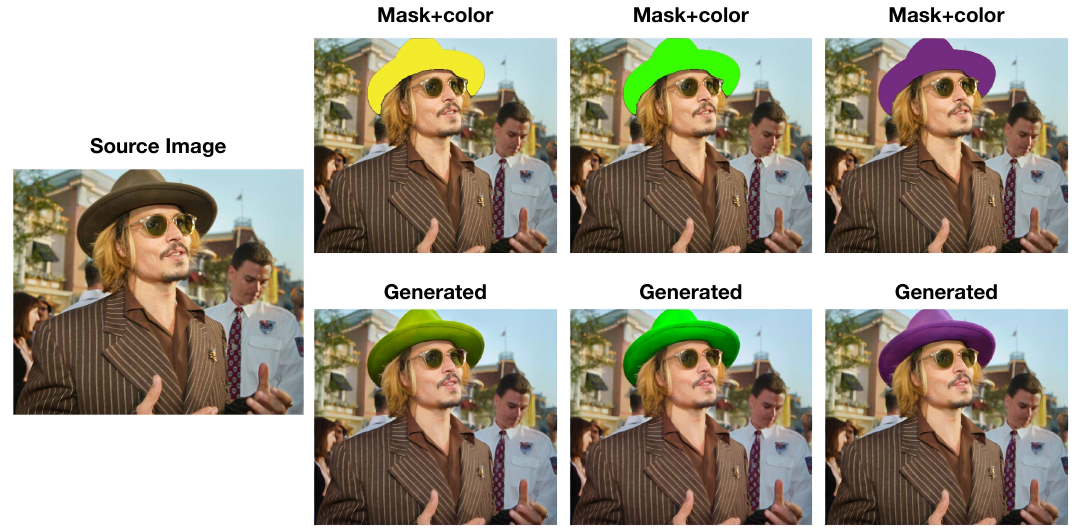}
\caption{Practical color-conditioned inpainting example. We obtain the sky (via selecting a Segment Anything patch~\cite{kirillov2023segment}), pick colors of interest, and have the conditional diffusion model dequantize (we use gradient-based conditioning here). The model harmonizes the result fairly well, respecting the requested color and mask.}
\label{fig:depp}
\end{figure*}

\begin{figure*}[ht]
\centering
\includegraphics[width=\linewidth]{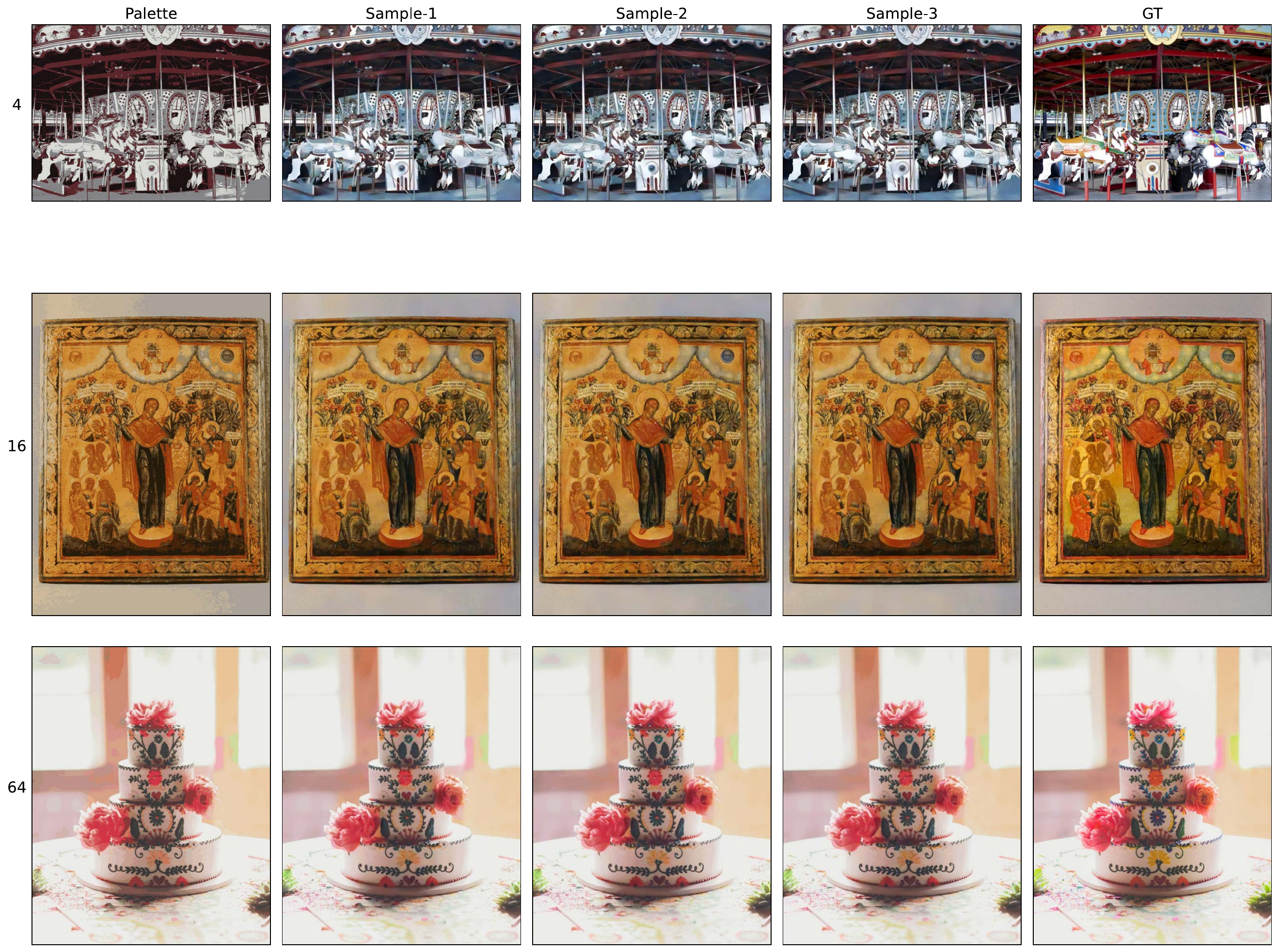}
\caption{Our image dequantizer can run inference at any aspect ratio, even though it was trained on square 256-res crops. Further, changing the rng during inference introduces some stochastic variety between samples. Samples \textbf{without texture conditioning} presented here, with number of colors in the palette specified in each row.}
\label{fig:stoch_no_tex}
\end{figure*}

\begin{figure*}[ht]
\centering
\includegraphics[width=\linewidth]{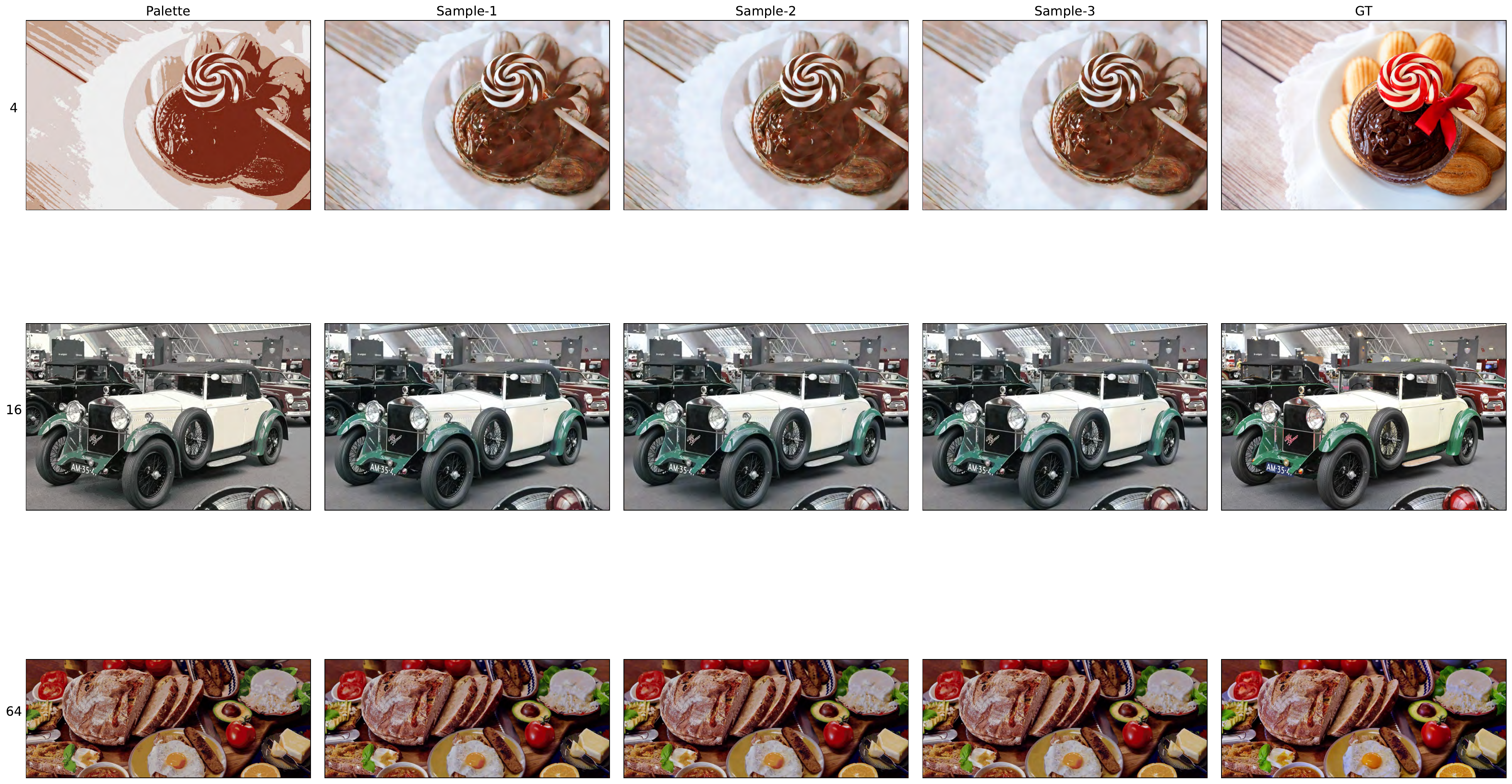}
\caption{Our image dequantizer can run inference at any aspect ratio, even though it was trained on square 256-res crops. Further, changing the rng during inference introduces some stochastic variety between samples. Samples \textbf{with texture conditioning} from $\mathsf{ours-T}$ presented here, with number of colors in the palette specified in each row.}
\label{fig:stoch_yes_tex}
\end{figure*}

\end{document}